\documentclass[10pt,twocolumn,letterpaper]{article}
\usepackage{geometry}
\newif\ifready
\readytrue
\ifready
    \usepackage{lib/cvpr}
\else
    \usepackage[review]{lib/cvpr}      %
\fi

\definecolor{TableBlue}{rgb}{0.17,0.49,0.75}
\definecolor{Cerulean}{rgb}{0,0,0.95}
\definecolor{LimeGreen}{rgb}{0.15,0.65,0.15}
\definecolor{RoyalBlue}{rgb}{0.25,0.41,0.88}
\definecolor{Rose}{rgb}{1.0, 0.15, 0.21}
\definecolor{Orange}{rgb}{1.0, 0.5, 0.0}
\definecolor{Gray}{gray}{0.6}
\definecolor{Black}{gray}{0.0}
\definecolor{Purple}{rgb}{0.77,0.12,0.64}

\newcommand{\change}[1]{{\color{Black}#1}}

\usepackage{mathrsfs}
\usepackage{times}
\usepackage{epsfig}
\usepackage{graphicx}
\usepackage{amsmath}
\usepackage{amssymb}
\usepackage{subcaption} %
\usepackage{graphicx,scalerel}

\usepackage{colortbl}
\usepackage{multirow}
\usepackage{hhline}

\usepackage{amsmath}

\usepackage{wrapfig}
\usepackage{soul}
\usepackage{siunitx}
\usepackage{multirow}
\usepackage{booktabs}
\usepackage{siunitx}
\usepackage{comment}

\definecolor{cvprblue}{rgb}{0.21,0.49,0.74}
\usepackage[pagebackref,breaklinks,colorlinks,allcolors=cvprblue]{hyperref}

\def\paperID{10820}
\def\confName{CVPR}
\def\confYear{2026}

\title{TeFlow: Enabling Multi-frame Supervision 
for Self-Supervised \\ Feed-forward Scene Flow Estimation}

\author{
Qingwen Zhang$^{1}$ 
\quad Chenhan Jiang$^{2}$
\quad Xiaomeng Zhu$^{1,5}$
\quad Yunqi Miao$^{3}$ \\
\quad Yushan Zhang$^{4}$
\quad Olov Andersson$^{1}$
\quad Patric Jensfelt$^{1}$ \\
\normalsize{$^{1}$KTH Royal Institute of Technology \quad $^{2}$Hong Kong University of Science and Technology} \\ 
\normalsize{$^{3}$University of Warwick \quad
$^{4}$Linköping University \quad $^{5}$Scania}
}

\begin{document}
\maketitle
\begin{abstract}
Self-supervised feed-forward methods for scene flow estimation offer real-time efficiency, but their supervision from two-frame point correspondences is unreliable and often breaks down under occlusions. Multi-frame supervision has the potential to provide more stable guidance by incorporating motion cues from past frames, yet naive extensions of two-frame objectives are ineffective because point correspondences vary abruptly across frames, producing inconsistent signals.
In the paper, we present TeFlow, enabling multi-frame supervision for feed-forward models by mining temporally consistent supervision. 
TeFlow introduces a temporal ensembling strategy that forms reliable supervisory signals by aggregating the most temporally consistent motion cues from a candidate pool built across multiple frames.
Extensive evaluations demonstrate that TeFlow establishes a new state-of-the-art for self-supervised feed-forward methods, achieving performance gains of \textbf{up to 33\%} on the challenging Argoverse 2 and nuScenes datasets. Our method performs on par with leading optimization-based methods, yet speeds up \textbf{150} times.
The code is open-sourced at \url{https://github.com/Kin-Zhang/TeFlow} along with trained model weights.
\end{abstract}

\vspace{-1.5em}
\section{Introduction}
\begin{figure}[t]
\centering
\includegraphics[width=0.95\linewidth]{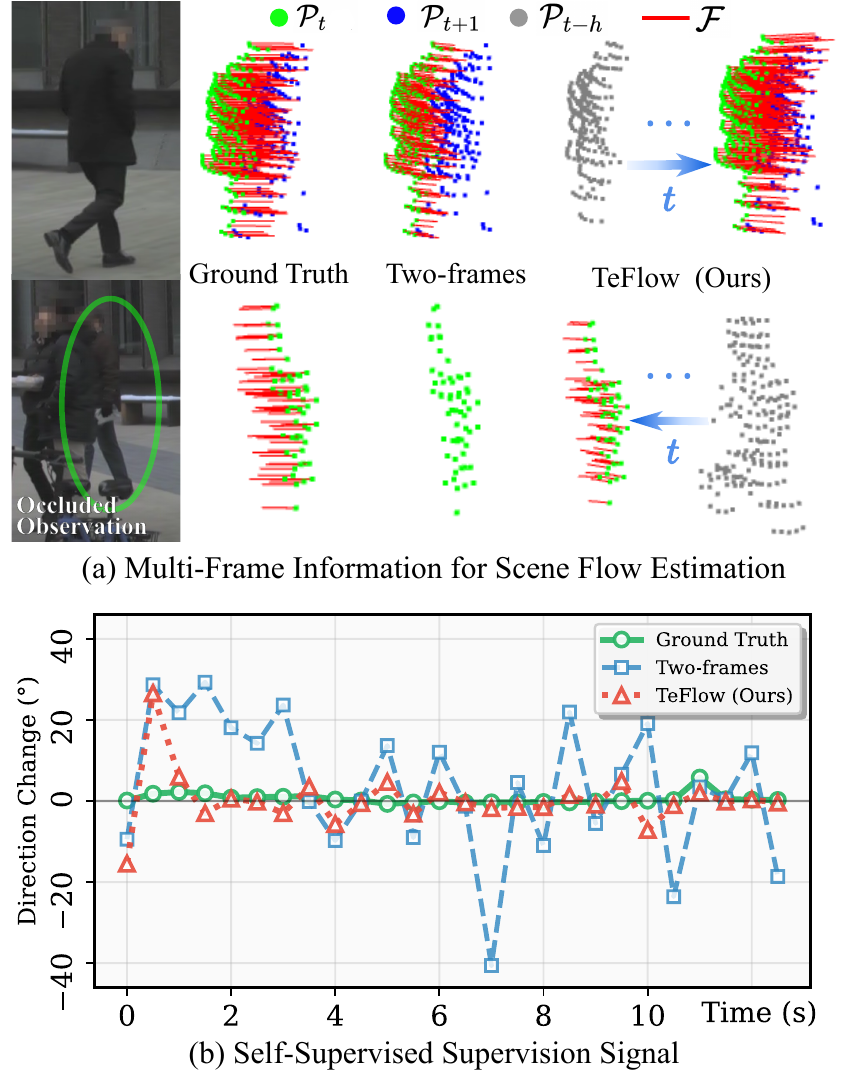}
\vspace{-1em}
\caption{(a) Multi-frame supervision maintains more stable guidance during occlusion by querying past frames, while two-frame supervision fails due to missing points. 
(b) Temporal consistency of supervision signals, measured as the change in motion direction (in degrees) between consecutive frames. 
The two-frame supervision~\cite{zhang2024seflow} exhibits abrupt variations with frequent direction shifts, while our five-frame TeFlow produces more stable signals that stay closer to the ground truth.
}
\vspace{-1.8em}
\label{fig:cover}
\end{figure}

\vspace{-0.5em}
Scene flow determines the 3D motion of each point between consecutive point clouds as visualized in~\Cref{fig:cover}a.
By providing a detailed characterization of object motion, scene flow could benefit downstream tasks such as motion prediction~\cite{najibi2022motion}, dynamic object reconstruction~\cite{chodosh2024smore,zhang2025himo}, and occupancy flow prediction~\cite{yang2024emernerf}.
Accurate scene flow prediction enables autonomous agents to capture the underlying environmental dynamics during observation~\cite{liso,zhang2023towards,li2025mos,jia2024beautymap}.

\begin{figure}[t]
\centering
\includegraphics[width=\linewidth]{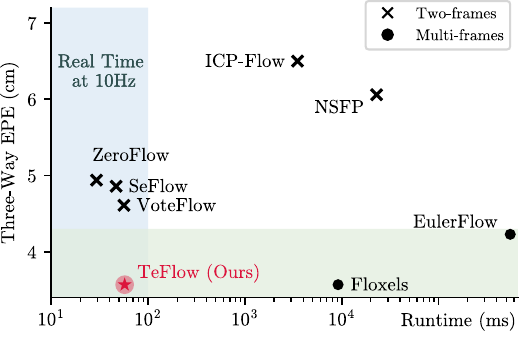}
\vspace{-3.2em}
\caption{Accuracy vs. Runtime. Prior feed-forward methods are fast but less accurate, while optimization-based methods are accurate but too slow. 
TeFlow achieves both real-time speed and high accuracy.}
\vspace{-2em}
\label{fig:cover_res}
\end{figure}

To overcome the high cost of manual annotation required by supervised methods~\cite{zhang2024deflow,fastflow3d,khoche2025ssf,mambaflow}, the field has increasingly shifted towards self-supervised learning, which exploits geometric and temporal consistency across frames without requiring ground-truth labels. 
Existing self-supervised approaches fall into two categories: 
(1)~Optimization-based methods~\cite{vedder2024neural,hoffmann2025floxels} achieve high accuracy by enforcing long-term multi-frame constraints but suffer from substantial optimization latency, making them unsuitable for real-time deployment. 
As shown in~\Cref{fig:cover_res}, the optimization of such methods can take hours and even days for a single scene. 
(2) Feed-forward methods~\cite{zhang2024seflow,lin2025voteflow} achieve high efficiency by generating results in a single forward pass, however, their accuracy is limited by unstable training objectives derived from only two consecutive frames. 
For example, as shown in~\Cref{fig:cover}a, when depicting objects (e.g., pedestrians), occlusions often cause missing points between frames, preventing consistent motion guidance and leading to incorrect flows. In addition, two-frame supervision is also vulnerable to sensor noise, sparse observations, and ambiguity in curved or articulated motion. Leveraging information from multiple frames mitigates these issues and provides a more stable and temporally consistent supervisory signal.

However, introducing additional frames into feed-forward training is non-trivial.
As shown in~\Cref{fig:cover}b, the direction of the two-frame supervisory signal varies drastically over time. Even when the underlying motion is smooth, two-frame estimates fluctuate sharply due to occlusions, noise, and missing points. Training with such temporally inconsistent signals prevents the model from learning coherent motion patterns and results in inaccurate scene flow. This highlights the importance of exploiting temporally consistent cues across multiple frames to provide effective self-supervision for feed-forward models.
To achieve this, we propose TeFlow, a novel multi-frame feed-forward framework that mines consistent motion signals across time. TeFlow introduces a temporal ensembling strategy that constructs a pool of motion candidates across multiple frames and applies a voting scheme to aggregate the most consistent ones. The resulting consensus motions form a robust supervisory signal, enabling feed-forward models to achieve high-accuracy scene flow estimation while maintaining real-time efficiency.
Our contributions can be summarized as follows:
\begin{itemize}
\item We leverage temporally-consistent supervisory signals for self-supervised scene flow estimation by constructing a motion candidate pool from multiple frames and then optimizing the consensus motion via a voting scheme.
\item By integrating our objective function, 
TeFlow becomes the first approach to unlock the potential of multi-frame architectures in a real-time, self-supervised setting.
\item We demonstrate through extensive experiments on Argoverse 2 and nuScenes datasets that TeFlow achieves the state-of-the-art performance for real-time self-supervised methods, significantly narrowing the accuracy gap to slow optimization-based methods while maintaining real-time efficiency across datasets.
\end{itemize}

\vspace{-4pt}
\section{Related Work}
\vspace{-4pt}
Scene flow estimation~\cite{vedula2005three,scoop,liudifflow3d2025,khatri2024can,jiang20243dsflabelling,zhang2024diffsf,ajinkya2026dogflow} has been a long-standing problem in computer vision. Our work builds upon advances in both supervised and, more importantly, self-supervised learning paradigms.

\vspace{2pt}
\noindent\textbf{Supervised Scene Flow.}
Early and many current state-of-the-art methods are trained in a fully supervised manner~\cite{wei2020pv,wang2023dpvraft,liu2024difflow3d,zhang2024gmsf,li2025uniflowzeroshotlidarscene}. These approaches leverage large datasets with ground-truth flow annotations to train deep neural networks. Methods like FastFlow3D~\cite{fastflow3d}, DeFlow~\cite{zhang2024deflow}, and SSF~\cite{khoche2025ssf} use voxel-based backbones to efficiently process large-scale point clouds, achieving high accuracy and real-time inference speeds. While powerful, these methods are fundamentally limited by their reliance on expensive, manually annotated data, which is difficult to scale and may not cover all real-world scenarios.
To overcome this limitation, in this work we focus on self-supervised scene flow estimation, which removes the need for labeled data.

\vspace{2pt}
\noindent\textbf{Self-Supervised Scene Flow.} 
Self-supervised methods eliminate the dependency on ground-truth annotations and can be broadly categorized into optimization-based and feed-forward approaches.

Optimization-based approaches fit a scene-specific model at test time. The pioneering NSFP~\cite{li2021neural} optimizes a small coordinate-based MLP for each two-frame pair. Follow-up works~\cite{li2023fast,hoffmann2025floxels} improve efficiency by replacing the MLP with representations like voxel grids or distance transforms. 
To achieve higher accuracy, the state-of-the-art method, EulerFlow~\cite{vedder2024neural}, reframes scene flow as the task of estimating a continuous ordinary differential equation over an entire sequence. By optimizing a neural prior against reconstruction objectives across many frames, it produces exceptionally accurate flow fields. However, this accuracy comes at a prohibitive computational cost, requiring from hours to days of optimization for a single sequence, making it unsuitable for any real-time application.

In contrast, feed-forward methods aim to train a single, generalizable network on a large unlabeled dataset, enabling real-time inference on new scenes. 
A prominent approach is knowledge distillation, exemplified by ZeroFlow~\cite{zeroflow}. This technique uses a slow but accurate optimization-based `teacher' to generate pseudo-labels for a fast `student' network. However, this label generation process requires 7.2 GPU months of computation, which limits its scalability and practical adoption.
Other methods, such as SeFlow~\cite{zhang2024seflow}, instead design the two-frame loss functions directly. SeFlow improves robustness by tailoring consistency losses based on static and dynamic classification, and mitigating overestimation through emphasizing larger nearest-neighbor displacements. However, despite these advancements, all existing feed-forward methods are fundamentally trained using supervisory signals derived from only two consecutive frames. This reliance makes the flow highly sensitive to occlusions and temporal inconsistency (\Cref{fig:cover}b). In contrast, we introduce TeFlow, the first self-supervised, real-time feed-forward approach to leverage multi-frame temporal consistency. By mining consistent motion cues across time, TeFlow produces a more stable supervisory signal, achieving high accuracy while preserving real-time performance.

\vspace{2pt}
\noindent\textbf{Multi-frame Architectures.}
Independent of the training paradigm, network architectures have evolved to better capture temporal information. Models like Flow4D~\cite{kim2024flow4d} introduce an explicit temporal dimension and use 4D convolutions~\cite{choy20194d} to process voxelized point-cloud sequences.
Taking a different approach to efficiency, DeltaFlow~\cite{zhang2025deltaflow} introduces a computationally lightweight `$\Delta$ scheme' that computes the difference between voxelized frames. This avoids the feature expansion common in other multi-frame methods and keeps the input size constant regardless of the number of frames. 
This architectural trend shows a clear recognition in the community that temporal context is crucial for accurate motion estimation. 
However, these multi-frame architectures have so far been effective mainly in supervised settings; under self-supervision, they remain constrained by two-frame objectives. TeFlow provides stable multi-frame supervision, allowing these architectures to operate at full temporal capacity in a self-supervised setting.

\vspace{-0.4em}
\section{Preliminaries}
\vspace{-0.4em}
\textbf{Problem Formulation.}
Given a continuous stream of LiDAR point clouds, our goal is to train a feed-forward network $\Phi_{\theta}$ that estimates the scene flow vector field~\cite{zhang2025deltaflow}. 
For a given frame $\mathcal{P}_{t} \in \mathbb{R}^{N_{t} \times 3}$, the network predicts its flow $\mathcal{F} \in \mathbb{R}^{N_{t} \times 3}$ toward the subsequent frame $\mathcal{P}_{t+1} \in \mathbb{R}^{N_{t+1} \times 3}$. 
The scene flow $\mathcal{F}$ is decomposed into two parts: ego-motion flow $\mathcal{F}_{\text{ego}}$ induced by the movement of the vehicle, and residual flow $\mathcal{F}_{\text{res}}$, caused by dynamic objects in the environment. Since ego-motion can be obtained directly from odometry, the network is trained only to estimate the residual flow. Formally, the network learns the mapping:
\begin{equation}
\Phi_{\theta}: \{ \mathbf{T}_{\text{ego}}^{t-h \rightarrow t+1}\mathcal{P}_{t-h}, \dots, \mathbf{T}_{\text{ego}}^{t \rightarrow t+1}\mathcal{P}_{t}, \mathcal{P}_{t+1} \} \rightarrow \mathcal{F}_{\text{res}},
\end{equation}
where $\mathbf{T}_{\text{ego}}^{t' \rightarrow t+1} \in \mathbb{R}^{4 \times 4}$ is the odometry transformation matrix from time $t'$ to $t+1$, aligning all past point clouds to the coordinate frame of $\mathcal{P}_{t+1}$.

\vspace{2pt}
\noindent\textbf{Self-Supervised Training Paradigm.}
To train $\Phi_{\theta}$ without labeled data, we adopt a self-supervised paradigm that derives supervisory signals directly from the input sequence. 
To enable targeted supervision for different motion patterns, point clouds are first segmented into static and dynamic regions $(\mathcal{P}_{\cdot,s}, \mathcal{P}_{\cdot,d})$. 
Static points $\mathcal{P}_{\cdot,s}$ are supervised with a near-zero flow loss.
Dynamic points $\mathcal{P}_{\cdot,d}$ are further partitioned into clusters $\mathcal{C} = \{\mathcal{C}_1, \mathcal{C}_2, \dots, \mathcal{C}_{N_c}\}$, where $N_c = |\mathcal{C}|$ is the number of dynamic clusters. 
Each cluster is assumed to undergo a shared rigid motion and is trained with a rigidity loss, i.e., dynamic cluster loss, enforcing coherent motion within the group.  This static–dynamic formulation has proven effective for modeling local motion coherence; however, prior works such as SeFlow~\cite{zhang2024seflow} were limited to two-frame supervision, making the learned signals susceptible to noise and temporal inconsistency. We instead aim to reformulate this concept toward stable, multi-frame training.

\begin{figure*}[t]
\centering
\includegraphics[trim=10 50 72 15, clip, width=\linewidth]{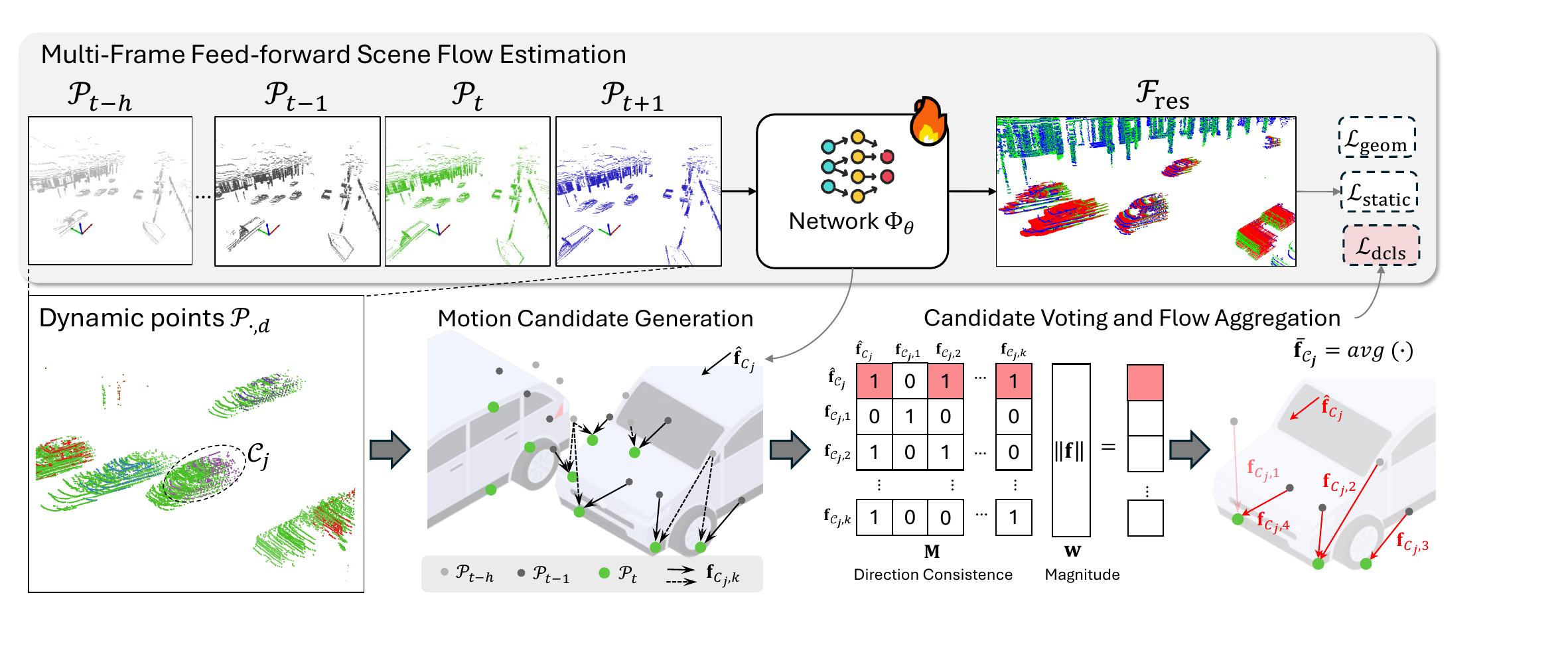}
\vspace{-1.5em}
\caption{\textbf{An overview of the TeFlow}, a multi-frame feedforward scene flow estimation pipeline, shown in the top row.
Our self-supervised pipeline tackles the main challenge of deriving reliable supervision $\bar{\textbf{f}}$ from dense multi-frame inputs. 
For each dynamic cluster $\mathcal{C}_j$, we constructs a motion candidate pool (internal $\hat{\textbf{f}}_{\mathcal{C}_j}$ and external $\mathbf f_{\mathcal C_j, k}$). Candidates are processed via a weighted consensus voting scheme using directional consistency $\mathbf{M}$ and magnitude-based reliability $\mathbf{w}$ to find a consensus winner (\Cref{eq:voting}). The final supervision $\bar{\mathbf{f}}_{\mathcal{C}_j}$ is a weighted average of the winner and agreeing candidates, which filters inconsistent outliers (e.g., $\hat{\textbf{f}}_{\mathcal{C}_j,1}$) for stable training.
}
\vspace{-1.2em}
\label{fig:framework}
\end{figure*}

\vspace{-4pt}
\section{Method: TeFlow}
\vspace{-4pt}
To move beyond the limits of two-frame supervision and achieve both high accuracy and efficiency, we propose TeFlow, a multi-frame feed-forward framework illustrated in~\Cref{fig:framework}, generating stable supervisory signals through temporal ensembling of consistent motions across frames.

\subsection{Temporal Ensembling for Dynamic Clusters}
\label{sec:tem_ensem}
TeFlow aims to assign each dynamic cluster $\mathcal{C}_j$ a reliable supervision target $\bar{\mathbf{f}}_{\mathcal{C}_j} \in \mathbb{R}^{1 \times 3}$ that reflects its true motion. 
A naive extension from two-frame to multi-frame supervision is unreliable, since frame-to-frame 
correspondences often vary abruptly and introduce conflicting signals (as shown in~\Cref{fig:cover}b). To address this, TeFlow introduces a temporal ensembling approach that first constructs a pool of motion candidates across the temporal window, capturing multiple hypotheses, and then forms a robust supervision signal by selecting and weighting the most consistent motions. 
The approach consists of two stages: (i) generate a diverse pool of motion candidates across the temporal window, and (ii) aggregate the target motion via a weighted voting scheme.

\vspace{2pt}
\noindent\textbf{Motion Candidate Generation.}
This stage aims to build a candidate pool from which a reliable supervisory target can be aggregated for each cluster $\mathcal{C}_j$. 
Each candidate is represented by a single 3D motion vector.
The pool combines two sources: internal and external candidates, which together balance stability with data-driven evidence.

The internal candidate $\hat{\mathbf{f}}_{\mathcal C_j}$ serves as an anchor that stabilizes the supervisory signal and keeps training grounded in the evolving state of the model. 
It is obtained from the current estimate from the network $\Phi_{\theta}$, computed as the average flow over all points in the cluster,
\vspace{-1em}

\begin{equation}
\hat{\mathbf{f}}_{\mathcal C_j} = \frac{1}{|\mathcal{C}_j|} \sum_{\mathbf{p}_i \in \mathcal{C}_j} \hat{\mathbf{f}}_{i},
\label{eq:internal_f}
\end{equation}

\vspace{-0.4em}\noindent where $|\mathcal{C}_j|$ is the number of point in the cluster and $\hat{\mathbf{f}}_{i} \in \mathcal F_{\text{res}}$ is the network estimation for point $\mathbf p_i$.

The external candidates $\mathbf{f}^{t'}_{\mathcal C_j,k}$ represent geometry-based motion hypotheses for the cluster. They aim to approximate how the cluster might actually move by exploiting information from neighboring frames. To construct them, we compare the cluster $\mathcal{C}_j$ at time $t$ with the dynamic points $\mathcal{P}_{t',d}$ from each of the other frames $t' \in \{ t-h, \dots, t-1, t+1 \}$. 
For every pair of frames $(t, t')$, we establish correspondences by finding, for each point $\mathbf{p}_i \in \mathcal{C}_j$, its nearest neighbor in $\mathcal{P}_{t',d}$. 
We then select the Top-$K$ correspondences with the largest displacement magnitudes, denoting their source points as $\mathbf{p}_k \in \mathcal{C}_j$. This selection helps capture meaningful motion while filtering out noise points.

Since different frames $t'$ are separated from $t$ by varying time intervals, the displacements are normalized by the temporal gap $(t'-t)$. 
The normalized motion vector for frame $t'$ and the $k$-th selected correspondence is defined as:
\begin{equation}
\mathbf{f}^{t'}_{\mathcal C_j,k} = \frac{\mathcal{NN}(\mathbf{p}_k, \mathcal{P}_{t',d}) - \mathbf{p}_k}{t' - t},
\label{eq:external_f}
\end{equation}

\vspace{-0.3em}
\noindent where $\mathcal{NN}(\cdot)$ denotes nearest-neighbor search.

Finally, we combine the internal candidate with all external candidates from the temporal window to form the complete candidate pool:
\begin{equation}
\resizebox{0.9\linewidth}{!}{$
\mathcal{F}_{\mathcal{C}_j} = \{\hat{\mathbf{f}}_{\mathcal{C}_j}\} \cup \{ \mathbf{f}^{t'}_{\mathcal{C}_j, k} \mid t' \in \{t-h, \dots, t-1,t+1\}, \ k \in\{1, \dots, K\} \}$}
\label{eq:full_pool}
\end{equation}

\vspace{-0.1em}
\noindent This pool contains a total of $1 + K(h+1)$ candidates, each candidate $\mathbf{f}_i \in \mathbb{R}^{1 \times 3}$. 
By uniting stability from the internal estimate with motion evidence from external correspondences, the pool provides a strong foundation for consensus in the subsequent voting stage.

\vspace{2pt}
\noindent\textbf{Candidate Voting and Flow Aggregation.}
With the candidate pool constructed, the next step is to derive a stable cluster-level flow. Since the pool still contains a mix of useful and noisy motion vectors, selecting one directly could lead to unstable supervision. 
To obtain a reliable estimate, we design a consensus-based voting scheme that aggregates candidates based on two criteria: (i) their agreement with others in the pool, and (ii) their own reliability.

The first criterion, agreement, captures which flows reinforce each other, ensuring that the final decision reflects collective support. 
It is measured through a consensus matrix $\mathbf{M} \in \mathbb{R}^{(1+K(h+1)) \times (1+K(h+1))}$. 
Each entry $\mathbf{M}_{ab}$ indicates whether two candidates $\mathbf{f}_a$ and $\mathbf{f}_b$ are directionally consistent, determined by their cosine similarity $\tau_{\text{cos}}$:
\vspace{-0.5em}

\begin{equation}
\mathbf{M}_{ab} =
\begin{cases}
1 & \text{if } \dfrac{\mathbf{f}_a \cdot \mathbf{f}_b}{\|\mathbf{f}_a\|\|\mathbf{f}_b\|} > \tau_{\text{cos}}, \\
0 & \text{otherwise}.
\end{cases}
\label{eq:direction}
\end{equation}

\vspace{-0.3em}\noindent The second criterion, reliability, reflects how trustworthy each candidate is and therefore how much influence it should have on the final flow. 
It is encoded in a weight vector $\mathbf{w} = [w_1, \dots, w_{1+K(h+1)}]^T$, where the weight of candidate $\mathbf{f}_i$ is defined as
\vspace{-0.5em}

\begin{equation}
w_i = \gamma^{m_i} \bigl(1 + \|\mathbf{f}_i\|^2_2\bigr).
\label{eq:rel_weight}
\end{equation}

\vspace{-0.3em}\noindent Here, $\gamma \in (0,1]$ is a temporal decay factor that prioritizes candidates from more recent frames, and $m_i$ is the time offset of $\mathbf{f}_i$, with $m_i = 0$ for the internal candidate and $m_i = |t'-t|$ for external ones. 
The magnitude term $\|\mathbf{f}_i\|^2_2$ further emphasizes larger displacements, which provide clearer motion cues than near-zero flows. 
This design encourages candidates with clearer motion cues to obtain higher weights and greater influence in the voting and aggregation stage.

With both agreement and reliability defined, we combine them to identify the most representative flow in the pool, referred to as the consensus winner. It is obtained as
\begin{equation}
a^\dagger = \underset{i \in \{1, \dots, 1+K(h+1)\}}{\arg\max} \, \mathbf{S}_i, \quad \text{where } \mathbf{S} = \mathbf{M}\mathbf{w}.
\label{eq:voting}
\end{equation}

\vspace{-0.3em}\noindent Here, each element \(\mathbf{S}_i\) aggregates the reliability weights of all candidates that agree with the \(i\)-th one, serving as the total vote score for candidate \(i\). A higher score means that a candidate is supported by more reliable neighbors. 
The index $a^\dagger$ therefore corresponds to the candidate with the strongest overall support.

Rather than relying only on this single winner, we further stabilize the supervision by taking a weighted average of flow candidates that are directionally consistent with the consensus winner:
\vspace{-0.8em}

\begin{equation}
\bar{\mathbf{f}}_{\mathcal C_j} = 
\frac{\sum_{b} \mathbf{M}_{a^{\dagger}b} w_b \mathbf{f}_b}{\sum_{b} \mathbf{M}_{a^{\dagger}b} w_b}.
\label{eq:avg_consesus_flow}
\end{equation}

\vspace{-0.2em}\noindent This averaging step preserves the reliability of the winner while incorporating supportive evidence from consistent candidates, thereby mitigating the effect of noise and producing a stable supervisory target from both model predictions and multi-frame geometric evidence.
As illustrated in~\Cref{fig:cover}b, this strategy yields supervisory signals that are significantly more consistent than those from two-frame supervision.
These signals $\bar{\mathbf{f}}_{\mathcal C_j}$ are then used to define our training objectives.

\subsection{Training Objective}
\label{sec:training_loss}
To train the model on dynamic objects, we define a dynamic cluster loss $\mathcal{L_{\text{dcls}}}$. Unlike prior two-frame methods~\cite{zhang2024seflow}, this loss is drived from our stable multi-frame supervision $\bar{\mathbf{f}}_{\mathcal C_j}$.  
The basic form is a \textit{point-level} L2 loss, computed between the model predictions and the supervisory targets and averaged over all points in all dynamic clusters.
However, as large objects contain more points, their losses dominate the training process, which biases the optimization and suppresses small objects.
To solve the problem, we introduce a \textit{cluster-level} loss term.
Specifically, this term first averages the L2 error within each cluster and then averages across clusters, ensuring that small objects contribute fairly rather than being overshadowed by larger ones. The proposed dynamic cluster loss $\mathcal{L}_{\text{dcls}}$ is the sum of the point-level and cluster-level terms:
\begin{equation}
\resizebox{0.9\linewidth}{!}{$
\mathcal{L}_{\text{dcls}} = 
\underbrace{ \frac{1}{|\mathcal P_{\mathcal C}|} \sum_{j} \sum_{\mathbf{p}_i \in \mathcal{C}_j} \| \hat{\mathbf{f}}_{i} - \bar{\mathbf{f}}_{\mathcal C_j} \|_2^2 }_{\text{Point-level Term}} 
+ \\
\underbrace{ \frac{1}{N_c} \sum_{j} \left( \frac{1}{|\mathcal{C}_j|} \sum_{\mathbf{p}_i \in \mathcal{C}_j} \| \hat{\mathbf{f}}_{i} - \bar{\mathbf{f}}_{\mathcal C_j} \|_2^2 \right) }_{\text{Cluster-level Term}},
\label{eq:our_dcls}$}
\end{equation}

\vspace{-0.3em}\noindent where $|\mathcal P_{\mathcal C}|$ is the total number of points across all dynamic clusters and $N_c$ is the number of clusters.  

In addition to our proposed $\mathcal{L}_{\text{dcls}}$, we adopt two auxiliary losses from prior work~\cite{zhang2024seflow,vedder2024neural} to ensure holistic training.
The \textit{static loss} $\mathcal{L}_{\text{static}}$~\cite{zhang2024seflow} penalizes non-zero residual flow on background points $\mathcal{P}_{t,s}$, since their motion is already explained by ego-motion. 
The \textit{geometric consistency loss} $\mathcal{L}_{\text{geom}}$ applies multi-frame Chamfer and dynamic Chamfer distances to ensure that the source point cloud, warped by the predicted flows, aligns with neighboring frames.

Together, these losses ensure that the network learns from reliable cluster-level supervision, respects static background constraints, and preserves global geometric consistency across time. The overall training objective is the sum of all three losses:
\begin{equation}
\mathcal{L}_{\text{total}} = \mathcal{L}_{\text{dcls}} + \mathcal{L}_{\text{static}} + \mathcal{L}_{\text{geom}}.
\label{eq:total_loss}
\end{equation}

\begin{table*}[t]
\setlength{\tabcolsep}{0.8em}
\centering
\caption{
Performance comparisons on the Argoverse 2 \underline{test set} leaderboard
~\cite{onlineleaderboard}. 
TeFlow achieves state-of-the-art performance in real-time scene flow estimation. 
`\#F' denotes the number of input frames. Runtime is reported per sequence (around 157 frames), with `-' indicating unreported values. 
Units are given in seconds (`s') and minutes (`m').
}
\label{tab:big_av2}
\vspace{-0.8em}
\resizebox{0.98\linewidth}{!}{
\begin{tabular}{lcc|cccc|ccccc} 
\toprule
\multicolumn{1}{c}{\multirow{2}{*}{Methods}} & \multirow{2}{*}{\#F} & \multirow{2}{*}{\begin{tabular}[c]{@{}c@{}}Runtime\\per seq\end{tabular}} & \multicolumn{4}{c|}{Three-way EPE (cm) ↓}                                                  & \multicolumn{5}{c}{Dynamic Bucket-Normalized ↓}                                                                   \\
\multicolumn{1}{c}{}                         &                      &                                                                           & Mean                 & FD                   & FS                   & BS                    & Mean                 & CAR                  & OTHER                & PED.                 & VRU                   \\ 
\midrule
Ego Motion Flow                              & -                    & -                                                                         & 18.13                & 53.35                & 1.03                 & 0.00                  & 1.000                & 1.000                & 1.000                & 1.000                & 1.000                 \\ 
\midrule
\textit{\footnotesize{Optimization-based}}                  & \multicolumn{1}{l}{} & \multicolumn{1}{l|}{}                                                     & \multicolumn{1}{l}{} & \multicolumn{1}{l}{} & \multicolumn{1}{l}{} & \multicolumn{1}{l|}{} & \multicolumn{1}{l}{} & \multicolumn{1}{l}{} & \multicolumn{1}{l}{} & \multicolumn{1}{l}{} & \multicolumn{1}{l}{}  \\
FastNSF~\cite{li2023fast}                                      & 2                    & 12m                                                                       & 11.18                & 16.34                & 8.14                 & 9.07                  & 0.383                & 0.296                & 0.413                & 0.500                & 0.322                 \\
NSFP~\cite{li2021neural}                                         & 2                    & 60m                                                                       & 6.06                 & 11.58                & 3.16                 & 3.44                  & 0.422                & 0.251                & 0.331                & 0.722                & 0.383                 \\
ICP-Flow~\cite{lin2024icp}                                     & 2                    & -                                                                         & 6.50                 & 13.69                & 3.32                 & 2.50                  & 0.331                & 0.195                & 0.331                & 0.435                & 0.363                 \\
Floxels~\cite{hoffmann2025floxels}                               & 13                   & 24m                                                                       & \textbf{3.57}        & 7.73                 & \textbf{1.44}        & \textbf{1.54}         & 0.154                & 0.112                & 0.213                & \textbf{0.195}       & 0.096                 \\
EulerFlow~\cite{vedder2024neural}                                    & all                  & 1440m                                                                     & 4.23                 & \textbf{4.98}        & 2.45                 & 5.25                  & \textbf{0.130}       & \textbf{0.093}       & \textbf{0.141}       & \textbf{0.195}       & \textbf{0.093}        \\ 
\midrule
\textit{\footnotesize{Feed-forward}}                        & \multicolumn{1}{l}{} & \multicolumn{1}{l|}{}                                                     & \multicolumn{1}{l}{} & \multicolumn{1}{l}{} & \multicolumn{1}{l}{} & \multicolumn{1}{l|}{} & \multicolumn{1}{l}{} & \multicolumn{1}{l}{} & \multicolumn{1}{l}{} & \multicolumn{1}{l}{} & \multicolumn{1}{l}{}  \\
ZeroFlow~\cite{zeroflow}                                     & 3                    & 5.4s                                                                      & 4.94                 & 11.77                & 1.74                 & 1.31                  & 0.439                & 0.238                & 0.258                & 0.808                & 0.452                 \\
SemanticFlow~\cite{chen2025semanticflow}                                 & 2                    & -                                                                         & 4.69                 & 12.26                & \textbf{1.41}        & \textbf{0.40}         & 0.331                & 0.210                & 0.310                & 0.524                & 0.279                 \\
SeFlow~\cite{zhang2024seflow}                                       & 2                    & 7.2s                                                                      & 4.86                 & 12.14                & 1.84                 & 0.60                  & 0.309                & 0.214                & 0.291                & 0.464                & 0.265                 \\
VoteFlow~\cite{lin2025voteflow}                                     & 2                    & 13s                                                                       & 4.61                 & 11.44                & 1.78                 & 0.60                  & 0.289                & 0.202                & 0.288                & 0.417                & 0.249                 \\
SeFlow++~\cite{zhang2025himo}                                     & 3                    & 10s                                                                       & 4.40                 & 10.99                & 1.44                 & 0.79                  & 0.264                & 0.209                & 0.272                & 0.367                & 0.210                 \\
TeFlow (Ours)                                & 5                    & 8s                                                                        & \textbf{3.57}        & \textbf{8.53}        & 1.49                 & 0.70                  & \textbf{0.205}       & \textbf{0.163}       & \textbf{0.227}       & \textbf{0.253}       & \textbf{0.177}        \\
\bottomrule
\end{tabular}
}
\vspace{-0.5em}
\end{table*}

\begin{table*}[t]
\setlength{\tabcolsep}{0.75em}
\centering
\caption{
Performance comparisons on the nuScenes \underline{validation set} with a 10Hz LiDAR frequency. 
TeFlow achieves state-of-the-art accuracy in scene flow estimation. 
Runtime is reported per sequence ($\approx$200 frames) using the same device.}
\label{tab:big_nus}
\vspace{-0.8em}
\resizebox{0.98\linewidth}{!}{
\begin{tabular}{lcc|cccc|ccccc} 
\toprule
\multicolumn{1}{c}{\multirow{2}{*}{Methods}} & \multirow{2}{*}{\#F} & \multirow{2}{*}{\begin{tabular}[c]{@{}c@{}}Runtime\\per seq\end{tabular}} & \multicolumn{4}{c|}{Three-way EPE (cm) ↓}                                                  & \multicolumn{5}{c}{Dynamic Bucket-Normalized ↓}                                                                   \\
\multicolumn{1}{c}{}                         &                      &                                                                           & Mean  & FD    & FS   & BS  & Mean   & CAR    & OTHER  & PED.   & VRU     \\ 
\midrule
Ego Motion Flow                              & -                    & -                                                                         & 12.34                & 35.94                & 1.07                 & 0.00        & 1.000                & 1.000                & 1.000                & 1.000                & 1.000                           \\ 
\midrule
\textit{\footnotesize{Optimization-based}}                           & \multicolumn{1}{l}{} & \multicolumn{1}{l|}{}                             & \multicolumn{1}{l}{} & \multicolumn{1}{l}{} & \multicolumn{1}{l}{} & \multicolumn{1}{l|}{} & \multicolumn{1}{l}{} & \multicolumn{1}{l}{} & \multicolumn{1}{l}{} & \multicolumn{1}{l}{} & \multicolumn{1}{l}{}  \\
NSFP~\cite{li2021neural}                                         & 2                    & 3.5m                                                                      & 10.79         & 20.26          & 4.88  & 7.23  & 0.602          & 0.463          & 0.456          & 0.829          & 0.662           \\
ICP-Flow~\cite{lin2024icp}                                      & 2                    & 3.2m                                                                     & 8.81          & 17.53          & 3.51  & 5.38  & 0.569          & 0.430          & 0.569          & 0.749          & 0.530          \\ 
FastNSF~\cite{li2023fast}                                      & 2                    & 2.6m                                                                      & 12.16         & 18.20          & 6.11  & 12.18 & 0.560          & 0.436          & 0.523          & 0.737          & 0.543           \\ 
\midrule
\textit{\footnotesize{Feed-forward}}                                 & \multicolumn{1}{l}{} & \multicolumn{1}{l|}{}                                                     & \multicolumn{1}{l}{} & \multicolumn{1}{l}{} & \multicolumn{1}{l}{} & \multicolumn{1}{l|}{} & \multicolumn{1}{l}{} & \multicolumn{1}{l}{} & \multicolumn{1}{l}{} & \multicolumn{1}{l}{} & \multicolumn{1}{l}{}  \\
SeFlow~\cite{zhang2024seflow}                                       & 2                    & 6s                                                                        & 8.19          & 16.15          & 3.97 & 4.45 & 0.544          & 0.396          & 0.635          & 0.726          & 0.419             \\
VoteFlow~\cite{lin2025voteflow}                                     & 2                    & 8s                                                                        & 7.80          & 15.65          & 3.51 & 4.24 & 0.538          & 0.355          & 0.605          & 0.780          & 0.410             \\
SeFlow++~\cite{zhang2025himo}                                     & 3                    & 7.5s                                                                      & 6.13          & 14.59          & 1.96 & 1.86 & 0.509          & 0.327          & 0.583          & 0.716          & 0.409             \\
TeFlow (Ours)                                & 5                    & 7s                                                                        & \textbf{4.64} & \textbf{10.92} & 1.49 & 1.51 & \textbf{0.395} & \textbf{0.303} & \textbf{0.461} & \textbf{0.474} & \textbf{0.344}    \\
\bottomrule
\end{tabular}
}
\vspace{-1.5em}
\end{table*}

\vspace{-8pt}
\subsection{Implementation Details}
\label{sec:imp_detail}
We build TeFlow on top of the multi-frame DeltaFlow backbone~\cite{zhang2025deltaflow}. Static and dynamic segmentation for training is provided by DUFOMap~\cite{daniel2024dufomap}, and dynamic clusters are pre-computed using HDBSCAN~\cite{campello2013density}.
The main hyperparameters of our method are as follows: a cosine similarity threshold of $\tau_{cos}=0.7071$ (corresponding to a $45^\circ$ angular difference), a Top-$K$ selection of $K=5$ for external candidates, and a temporal decay factor of $\gamma=0.9$. For the DeltaFlow backbone, we adopt its standard configuration, processing a $76.8 \times 76.8 \times 6$\,m region represented as a $512 \times 512 \times 40$ voxel grid with $0.15$\,m resolution.
Training is performed for 15 epochs using the Adam optimizer~\cite{loshchilov2017fixing} with a learning rate of $0.002$ and a total batch size of 20, distributed across ten NVIDIA RTX 3080 GPUs. Each dataset requires approximately 15 to 20 hours of training. 
\ifready
    The implementation and pretrained weights are publicly available at \url{https://github.com/Kin-Zhang/TeFlow}.
\else
    The source code and model weights will be released upon publication.
\fi

\vspace{-4pt}
\section{Experiments}
\vspace{-4pt}
\textbf{Datasets.}
Experiments are conducted on two large-scale autonomous driving datasets: Argoverse 2~\cite{Argoverse2_2021}, collected with two roof-mounted 32-channel LiDARs, and nuScenes~\cite{nuscenes}, which uses a single 32-channel LiDAR. Details on datasets description, preprocessing, and ground-truth flow estimation are provided in~\Cref{app:dataset}.
We additionally evaluate TeFlow on the Waymo Open Dataset~\cite{sun2020scalability}, with results and analysis presented in~\Cref{app:more_study}.

\vspace{2pt}
\noindent\textbf{Evaluation Metrics.}
We follow the official Argoverse 2 benchmark and report three-way End Point Error (EPE)~\cite{Chodosh_2024_WACV} and Dynamic Bucket-Normalized EPE~\cite{khatri2024can}. 
EPE measures the L2 distance between predicted and ground-truth flow vectors in meters.
\textit{Three-way EPE} computes the unweighted average EPE over three categories: foreground dynamic (FD), foreground static (FS), and background static (BS).
\textit{Dynamic Bucket-Normalized EPE} normalizes the EPE by the mean speed within predefined motion buckets, providing a fairer comparison across different object classes. It evaluates four categories: regular cars (CAR), other vehicles (OTHER), pedestrians (PED.), and wheeled vulnerable road users (VRU).
All evaluations are conducted within a 70$\times$70 m area around the ego vehicle.

\noindent\textbf{Baselines.}
We compare TeFlow against both optimization-based and feed-forward self-supervised methods: NSFP~\cite{li2021neural}, FastNSF~\cite{li2023fast}, ICPFlow~\cite{lin2024icp}, ZeroFlow~\cite{zeroflow}, SeFlow~\cite{zhang2024seflow}, SemanticFlow~\cite{chen2025semanticflow}, SeFlow++~\cite{zhang2025himo}, EulerFlow~\cite{vedder2024neural}, VoteFlow~\cite{lin2025voteflow} and Floxels~\cite{hoffmann2025floxels}. 
To ensure fairness, Argoverse 2 results are obtained directly from the public leaderboard, and nuScenes baselines are reproduced following OpenSceneFlow\footnote{\url{https://github.com/KTH-RPL/OpenSceneFlow}},
using the best reported training configurations.

\begin{table*}[t]
\centering
\setlength{\tabcolsep}{0.8em}
\caption{
Ablation on the number of input frames on the Argoverse 2 validation set. 
All experiments use the same DeltaFlow backbone for a fair comparison. 
TeFlow surpasses SeFlow even with two frames, and performance peaks at five frames, indicating the optimal temporal window.
The best results are shown in \textbf{bold}.
}
\label{tab:frame}
\vspace{-0.8em}
\small
\resizebox{0.91\linewidth}{!}{
\begin{tabular}{cc|ccccc|cccc} 
\toprule
\multirow{2}{*}{Loss Type} & \multirow{2}{*}{\#Frame} & \multicolumn{5}{c|}{Dynamic Bucket-Normalized ↓}                                   & \multicolumn{4}{c}{Three-way EPE (cm) ↓}      \\
                           &                      & Mean           & CAR            & OTHER         & PED.           & VRU         & Mean          & FD             & FS   & BS    \\ 
\midrule
SeFlow                     & 2                    & 0.408          & 0.319          & 0.412          & 0.369          & 0.531          & 6.35          & 16.63          & 1.48 & 0.92  \\ 
\midrule
\multirow{5}{*}{TeFlow}    & 2                    & 0.353          & 0.271          & 0.389          & 0.329          & 0.424          & 5.98          & 13.93          & 2.53 & 1.46  \\
                           & 4                        & 0.283 & 0.204 & 0.342  & 0.295 & 0.293           & 4.57 & 10.77 & 1.87 & 1.08                \\
                           & 5                    & \textbf{0.265} & \textbf{0.198} & \textbf{0.275} & 0.295          & \textbf{0.293} & \textbf{4.43} & \textbf{10.36} & 1.86 & 1.08  \\
                           & 6                        & 0.269 & 0.197 & 0.305  & 0.290 & 0.284           & 4.55 & 10.66 & 1.87 & 1.12                \\
                           & 8                    & 0.300          & 0.269          & 0.336          & \textbf{0.273} & 0.321          & 5.40          & 13.50          & 1.78 & 0.91  \\
\bottomrule
\end{tabular}}
\vspace{-0.8em}
\end{table*}

\begin{table*}[h]
\centering
\setlength{\tabcolsep}{0.8em}
\caption{
Ablation study of proposed self-supervised loss items (See~\Cref{eq:total_loss}). Results are evaluated on the Argoverse 2 validation set using default hyperparameters.
\textbf{Bold} indicates the best performance and {\color{purple}purple} highlights settings with a significant performance drop.
}
\vspace{-0.8em}
\label{tab:fullabloss}
\resizebox{0.91\linewidth}{!}{
\begin{tabular}{lcc|ccccc|cccc} 
\toprule
\multicolumn{3}{c|}{Loss item} & \multicolumn{5}{c|}{Dynamic Bucket-Normalized ↓}                                                                                   & \multicolumn{4}{c}{Three-way EPE (cm) ↓}                       \\
\(\mathcal{L}_{\text{gemo}}\) & \(\mathcal{L}_{\text{static}}\) & \(\mathcal{L}_{\text{dcls}}\)   & Mean            & CAR             & OTHER                              & PED.                       & VRU                        & Mean          & FD            & FS            & BS             \\ 
\midrule
\checkmark      &             &                     & 0.386 & 0.317 & 0.586  & 0.297 & 0.343          & 8.85 & 17.26 & 4.45 & 4.85 \\
\checkmark      & \checkmark  &                     & \textcolor{purple}{0.458}             & 0.321            & \textcolor{purple}{0.654}   & 0.481            & 0.377   & 6.37          & 17.15          & \textbf{1.25}                    & \textbf{0.73}    \\ 
                &             & \checkmark          & 0.303             & 0.254            & 0.310   & \textbf{0.285}            & 0.362   & 8.53          & 12.28          & \textcolor{purple}{7.17}   & \textcolor{purple}{6.14}    \\ 
                & \checkmark  & \checkmark          & 0.313 & 0.233 & 0.402  & 0.296 & 0.321          & 4.84 & 11.99 & 1.73 & 0.80                \\
\checkmark      & \checkmark  & \checkmark          & \textbf{0.265}    & \textbf{0.198}   & \textbf{0.275}   & 0.295   & \textbf{0.293}   & \textbf{4.43} & \textbf{10.36} & 1.86        & 1.08           \\
\bottomrule
\end{tabular}
}
\vspace{-1.4em}
\end{table*}

\begin{table}[h]
\centering
\caption{
Ablation study of our dynamic cluster loss (See~\Cref{eq:our_dcls}). Results are evaluated on the Argoverse 2 validation set.
}
\vspace{-0.8em}
\label{tab:dcls_term}
\resizebox{\linewidth}{!}{
\begin{tabular}{l|ccccc} 
\toprule
\multicolumn{1}{c|}{\multirow{2}{*}{Formulation}} & \multicolumn{5}{c}{Dynamic Bucket-Normalized ↓} \\
\multicolumn{1}{c|}{}                                 & Mean           & CAR            & OTHER         & PED.           & VRU        \\ 
\midrule
Point-level         & 0.351 & 0.258 & 0.331 & 0.352 & \textcolor{purple}{0.463} \\
Cluster-level       &  0.356 & 0.222 & \textcolor{purple}{0.603} & 0.284 & 0.316 \\
$\mathcal L_{\text{dcls}}$ (Ours)     & \textbf{0.265} & \textbf{0.198} & \textbf{0.275} & \textbf{0.295} & \textbf{0.293} \\
\bottomrule
\end{tabular}}
\vspace{-1.8em}
\end{table}

\vspace{-4pt}
\subsection{State-of-the-art Comparison}
\vspace{-4pt}
TeFlow achieves state-of-the-art accuracy on both Argoverse 2 and nuScenes while maintaining real-time efficiency, as shown in~\Cref{tab:big_av2} and~\Cref{tab:big_nus} respectively. On Argoverse 2 test set, TeFlow achieves a Three-way EPE of 3.57 cm, on par with the best optimization-based method Floxels, while being 150× faster (8 s vs 24 min). On Dynamic Bucket-Normalized EPE, TeFlow improves by 22.3\% overall compared to SeFlow++, with consistent gains across all categories, including a 31\% error reduction for pedestrians. On nuScenes validation set, TeFlow again outperforms all baselines. 
It achieves the best dynamic normalized score (0.395) and the lowest Three-way EPE (4.64 cm), representing a 22.4\% improvement over SeFlow++. The most significant advance is the 33.8\% error reduction for the challenging pedestrian class.
TeFlow also achieves state-of-the-art performance on the Waymo dataset (See~\Cref{app:more_study}).
Together, these results show that TeFlow delivers optimization-level accuracy while retaining the efficiency and scalability of feed-forward methods, setting a new state-of-the-art for self-supervised scene flow estimation. 

\vspace{-4pt}
\subsection{Ablation Studies}
\vspace{-4pt}
\label{sec:ab_num_frames}
To further understand the source of performance gains in TeFlow, we conduct ablation studies on Argoverse 2. Additional analyses (e.g., backbone analysis, hyperparameter selection) are in \Cref{app:more_study}.

\vspace{1pt}
\noindent\textbf{Number of Input Frames.}
\Cref{tab:frame} ablates the impact of the number of input frames. 
\textit{Two-frame setting:} 
To assess the contribution of our formulation, we re-implement SeFlow on the same DeltaFlow backbone with an identical two-frame input. TeFlow achieves a 13.5\% reduction in dynamic EPE (0.353 vs. 0.408), mainly due to our candidate pool and the cluster-level dynamic loss term, which provides more consensus information and ensures balanced supervision across object sizes.
\textit{Multi-frame setting:} Expanding the temporal window within TeFlow to five frames yields the best performance, lowering dynamic EPE by 24.9\% to 0.265. 
This performance gain can be explained by~\Cref{fig:cover}b: the multi-frame supervision produced by TeFlow closely follows the ground truth and is more stable than the fluctuating signals from two-frame supervision. Training with these stable signals results in significantly better performance.
Further extending the number of frames shows little help or even degrades the performance, which is consistent with the prior findings in the supervised method~\cite{kim2024flow4d,zhang2025deltaflow}.
A possible explanation is that overly distant frames introduce noisy or less relevant motion, outweighing the benefits of a longer context.

\begin{figure*}[t]
\centering
\includegraphics[trim=0 220 0 20, clip, width=\linewidth]{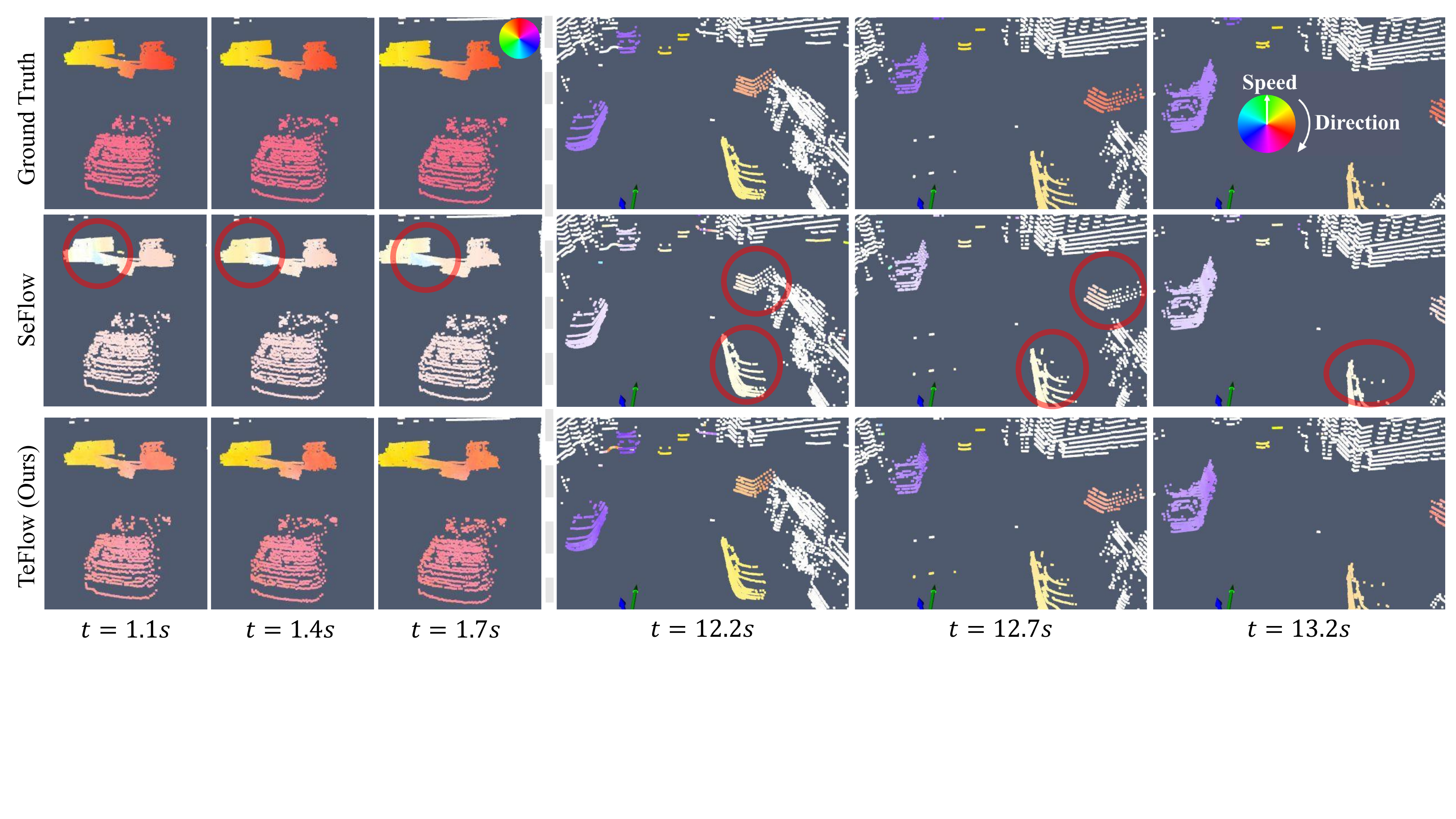}
\vspace{-1.8em}
\caption{Qualitative results on Argoverse 2 (left) and nuScenes (right). Rows show ground truth, SeFlow, and TeFlow predictions across time. 
Scene flow is visualized with hue indicating direction and saturation representing speed. 
Compared to SeFlow, TeFlow produces flow estimates that are more accurate and temporally consistent, particularly for dynamic objects (red circles). 
}
\vspace{-1.5em}
\label{fig:qual}
\end{figure*}

\vspace{-1pt}
\noindent\textbf{Self-supervised Loss Item.}
\Cref{tab:fullabloss} evaluates the contribution of each loss term in our proposed self-supervised objective.
Using only the geometric loss provides limited supervision, as nearest-neighbor alignment provides coarse motion cues.
Adding the static term improves three-way EPE but increases the dynamic normalized error.
Training with only the proposed dynamic-cluster loss \(\mathcal{L}_{\text{dcls}}\) achieves strong dynamic performance, especially for pedestrians, since the temporal ensembling discovers reliable supervision from multi-frame consistency; however, the absence of static constraints leads to large errors in static regions (FS, BS).
Combining \(\mathcal{L}_{\text{static}}\) with \(\mathcal{L}_{\text{dcls}}\) restores balanced accuracy, while incorporating all three losses delivers the best result (0.265), reducing the dynamic normalized error by 31.3\% compared to the geometric baseline and demonstrating that our multi-frame self-supervised objective effectively unifies geometric, static, and dynamic cues into a consistent training signal.

\vspace{-1pt}
\noindent\textbf{Dynamic Cluster Loss Formulation.}
\Cref{tab:dcls_term} evaluates the contribution of the point-level and cluster-level terms in the proposed dynamic cluster loss $\mathcal L_{\text{dcls}}$ (\Cref{eq:our_dcls}).
Training with only the point-level term underperforms on small and slow-moving agents such as pedestrians and VRUs, supporting our claim that point-wise supervision is dominated by large clusters containing many points.
When trained with only the cluster-level term, the model improves small-object performance but loses fine-grained alignment for large dynamic objects, resulting in an 82\% increase in the OTHER category error (0.603 vs.\ 0.331).
Combining both terms achieves the best overall performance, reducing the dynamic normalized error by 24.5\% and 25.6\% compared to the point-level and cluster-level variants, respectively, demonstrating the effectiveness of our proposed self-supervised formulation in providing reliable and balanced supervision across different object scales.
\begin{table}[t]
\centering
\caption{
Ablation study of our candidate pool (See~\Cref{eq:full_pool}). Results are evaluated on the Argoverse 2 validation set.
}
\vspace{-0.8em}
\label{tab:candidate_pool}
\resizebox{\linewidth}{!}{
\begin{tabular}{l|ccccc} 
\toprule
\multicolumn{1}{c|}{\multirow{2}{*}{Candidates Pool}} & \multicolumn{5}{c}{Dynamic Bucket-Normalized ↓}\\
\multicolumn{1}{c|}{}                                 & Mean           & CAR            & OTHER         & PED.           & VRU           \\ 
\midrule
Internal Only                                     & 0.455 & 0.232 & 0.481 & 0.505 & 0.601 \\
External Only                                     & 0.321 & 0.278 & 0.403 & \textbf{0.281} & 0.321 \\
Both (Proposed)                                       & \textbf{0.265} & \textbf{0.198} & \textbf{0.275} & 0.295 & \textbf{0.293}  \\
\bottomrule
\end{tabular}}
\vspace{-1.9em}
\end{table}

\noindent\textbf{Candidate Pool.}
\Cref{tab:candidate_pool} evaluates how different candidate sources contribute to our multi-frame self-supervision pipeline.  
The internal-only variant, which removes geometric cues entirely, gives the weakest result and shows that network predictions alone cannot provide reliable motion evidence.  
The external-only variant improves performance by incorporating multi-frame geometric cues, notably excels on small dynamic objects like PED. (44.4\% decreased), yet it yields less accurate supervision on objects with more stable motion directions such as CAR (19.8\% increased) due to the lack of a stabilizing internal reference. 
Using both candidates achieves the best overall accuracy, confirming that the internal candidate serves as a stable anchor while external candidates provide complementary motion hypotheses, resulting in a 41.8\% and 17.6\% error reduction over internal and external only methods, respectively.

\vspace{-0.4em}
\subsection{Qualitative results}
\vspace{-3pt}
\Cref{fig:qual} presents qualitative comparisons of two challenging dynamic scenarios. On the left (Argoverse 2), the scene contains a moving car and an articulated truck making a turn. The cab and trailer of the truck exhibit distinct motions that are visible in the ground truth, but SeFlow fails to capture them and predicts a uniform flow across the entire vehicle. TeFlow, in contrast, models the articulated components more accurately, producing flows that closely match the ground truth. On the right (nuScenes), a multi-object scene is shown. Here, estimates for the moving vehicle (red circles) in SeFlow are unstable and flicker across frames, while TeFlow delivers stable and temporally consistent flow throughout the sequence. More qualitative examples are provided in~\Cref{app:qual}.

\vspace{-8pt}
\section{Conclusion}
\vspace{-6pt}
In this work, we introduced TeFlow, a self-supervised feed-forward approach that unlocks the benefits of multi-frame supervision for real-time scene flow estimation. By mining temporally consistent supervisory signals through our temporal ensembling and voting strategy, TeFlow overcomes the limitations of traditional two-frame supervision and unstable point-wise correspondences.
On the Argoverse 2 and nuScenes benchmarks, TeFlow sets a new state-of-the-art for self-supervised, real-time methods, improving accuracy by up to 33\%. It successfully closes the gap with slower optimization-based approaches, offering comparable performance at a 150x speedup, thereby achieving both high accuracy and efficiency.

\ifready
\paragraph{Acknowledgement}
This work was partially supported by the Wallenberg AI, Autonomous Systems and Software Program (WASP) funded by the Knut and Alice Wallenberg Foundation. 
The computations and data handling were enabled by the Berzelius supercomputing resource provided by the National Supercomputer Centre at Linköping University and the Knut and Alice Wallenberg Foundation, Sweden, as well as by resources provided by Chalmers e-Commons at Chalmers and the National Academic Infrastructure for Supercomputing in Sweden (NAISS), partially funded by the Swedish Research Council through grant agreement no. 2022-06725.

\fi
{
    \small
    \bibliographystyle{lib/ieeenat_fullname}
    \bibliography{ref}
}
\clearpage
\clearpage
\newpage
\setlength\parindent{0pt}
\newgeometry{margin=2.54cm}
\onecolumn
\appendix
\vspace{1em}
\section*{\centering \Large TeFlow: Enabling Multi-frame Supervision for Self-Supervised Feed-forward Scene Flow Estimation \\ 
\vspace{4pt}
\normalsize{Supplementary Material}
}
\vspace{1em}
\section{Datasets Description}
\label{app:dataset}
We evaluate our method on three major autonomous driving benchmarks: Argoverse 2, nuScenes, and Waymo Dataset.

The Argoverse 2 dataset is a primary benchmark for scene flow estimation, featuring 700 training, 150 validation, and 150 test scenes, totaling approximately 107,000 training frames. 
Our main evaluations are conducted on the official test split, with results compared against the Argoverse 2 Scene Flow Challenge leaderboard~\cite{khatri2024can}, which provides official baseline results.
For the local validation, we follow~\cite{zhang2025deltaflow} and apply dynamic motion compensation to generate ground-truth flow labels.

The nuScenes dataset~\cite{nuscenes} contains 700 training and 150 validation scenes. Since nuScenes does not provide official scene flow annotations, we follow the protocol of~\cite{zhang2025himo} to generate pseudo ground truth. To ensure a consistent temporal resolution, the native 20Hz LiDAR data is first downsampled to 10Hz, resulting in a standard 100ms interval between frames. For each object, a rigid transformation is estimated from its 3D bounding box annotations and instance ID. This transformation is then applied to all LiDAR points within the object to compute their displacements, which serve as pseudo ground-truth flow labels. These labels are generated only for the validation set for evaluation, while training uses the full 137,575 unlabeled frames, demonstrating the scalability of our self-supervised approach.

The Waymo dataset~\cite{fastflow3d,sun2020scalability} is captured by a custom 64-channel LiDAR and contains 798 training and 202 validation sequences, each recorded at 10$\mathrm{~Hz}$ for around 20 seconds. The training set consists of 155,000 frames.

For all datasets, ground points are removed prior to evaluation. In Argoverse 2 and Waymo, we use the provided HD maps following the official protocol, whereas in nuScenes we apply a line-fitting-based ground segmentation method~\cite{himmelsbach2010fast}. As a result, all reported evaluation metrics are computed exclusively on non-ground points.

\section{Additional Quantitative Analysis}
\label{app:more_study}

\paragraph{State-of-the-art Comparison in Waymo}
Following our extensive evaluation on Argoverse 2 and nuScenes in the main text, we further validate TeFlow on the Waymo dataset to demonstrate the method's robustness across diverse scenarios and sensor configurations.
As shown in~\Cref{tab:big_waymo}, compared to the second-best feed-forward baseline (SeFlow++), TeFlow achieves a 14.9\% reduction in Dynamic Bucket-Normalized EPE (0.275 vs. 0.323), with consistent improvements across all object categories. 
Notably, the error for VRU (e.g., cyclist) is reduced by 19.0\% (0.198 vs. 0.247), highlighting the benefit of reliable multi-frame temporal modeling for tracking small, fast-moving agents.

In terms of absolute accuracy on the standard Three-way EPE metric, TeFlow maintains its lead over all baselines. 
It attains a mean EPE of 3.48 cm, surpassing SeFlow++ (3.63 cm), while achieving the lowest dynamic EPE (8.61 cm), significantly outperforming both feed-forward and per-scene optimization-based methods like NSFP (17.12 cm) and ICP-Flow (20.81 cm). 
These consistent results across three major datasets confirm that TeFlow establishes a significantly more effective self-supervised learning framework, capable of mining reliable multi-frame supervision across diverse scenarios.

\begin{table*}[h]
\centering
\setlength{\tabcolsep}{0.9em}
\caption{
Performance comparisons on the Waymo \underline{validation set}. 
TeFlow achieves state-of-the-art accuracy in scene flow estimation. 
Runtime is reported per sequence ($\approx$200 frames) using the same device. The best results are shown in \textbf{bold}.}
\label{tab:big_nus}
\vspace{-0.8em}
\resizebox{0.92\linewidth}{!}{
\begin{tabular}{lcccccc|cccc} 
\toprule
\multicolumn{1}{c}{\multirow{2}{*}{Methods}} & \multirow{2}{*}{\#F} & \multirow{2}{*}{\begin{tabular}[c]{@{}c@{}}Runtime\\per seq\end{tabular}} & \multicolumn{4}{c|}{Dynamic Bucket-Normalized ↓}                                           & \multicolumn{4}{c}{Three-way EPE (cm) ↓}                                                   \\
\multicolumn{1}{c}{}                         &                      &                                                                           & Mean                 & CAR                  & PED                  & VRU                   & Mean                 & FD                   & FS                   & BS                    \\ 
\midrule
Ego Motion Flow                              & -                    & -                                                                         & 1.000                & 1.000                & 1.000                & 1.000                 & 17.10                & 50.85                & 0.46                 & 0.00                  \\ 
\midrule
\textit{\footnotesize{Optimization-based}}                           & \multicolumn{1}{l}{} & \multicolumn{1}{l}{}                                                      & \multicolumn{1}{l}{} & \multicolumn{1}{l}{} & \multicolumn{1}{l}{} & \multicolumn{1}{l|}{} & \multicolumn{1}{l}{} & \multicolumn{1}{l}{} & \multicolumn{1}{l}{} & \multicolumn{1}{l}{}  \\
NSFP~\cite{li2021neural}                                         & 2                    & 96m                                                                       & 0.574                & 0.315                & 0.823                & 0.584                 & 10.05                & 17.12                & 10.81                & 2.21                  \\
ICP-Flow~\cite{lin2024icp}                                     & 2                    & 11m                                                                       & 0.328                & 0.305                & 0.485                & 0.195                 & 8.50                 & 20.81                & 2.14                 & 2.57                  \\
FastNSF~\cite{li2023fast}                                      & 2                    & 6.7m                                                                      & 0.458                & 0.236                & 0.719                & 0.418                 & 9.24                 & 18.19                & 2.56                 & 6.98                  \\ 
\midrule
\textit{\footnotesize{Feed-forward}}                                 & \multicolumn{1}{l}{} & \multicolumn{1}{l}{}                                                      & \multicolumn{1}{l}{} & \multicolumn{1}{l}{} & \multicolumn{1}{l}{} & \multicolumn{1}{l|}{} & \multicolumn{1}{l}{} & \multicolumn{1}{l}{} & \multicolumn{1}{l}{} & \multicolumn{1}{l}{}  \\
ZeroFlow~\cite{zeroflow}                                     & 2                    & 11.5s                                                                     & 0.770                & 0.444                & 0.982                & 0.884                 & 8.64                 & 22.40                & 1.57                 & 1.96                  \\
SeFlow~\cite{zhang2024seflow}                                       & 2                    & 12s                                                                       & 0.351                & 0.212                & 0.551                & 0.289                 & 4.29                 & 10.49                & 1.39                 & 1.00                  \\
VoteFlow~\cite{lin2025voteflow}                                     & 2                    & 13.5s                                                                     & 0.347                & 0.197                & 0.548                & 0.298                 & 3.89                 & 9.65                 & 1.12                 & 0.88                  \\
SeFlow++~\cite{zhang2025himo}                                     & 3                    & 16s                                                                       & 0.323                & 0.201                & 0.521                & 0.247                 & 3.63                 & 9.30                 & \textbf{0.87}        & \textbf{0.71}         \\
TeFlow (Ours)                                & 5                    & 14s                                                                       & \textbf{0.275}       & \textbf{0.157}       & \textbf{0.469}       & \textbf{0.198}        & \textbf{3.48}        & \textbf{8.61}        & 0.97                 & 0.86                  \\
\bottomrule
\end{tabular}}
\label{tab:big_waymo}
\vspace{-1em}
\end{table*}

\paragraph{Different Multi-frame Backbone}
\Cref{tab:diff_backbone} evaluates the generality of our self-supervised framework across different multi-frame backbones.
We first adapt Flow4D, an architecture originally designed for supervised learning, to the self-supervised setting. Enabled by TeFlow's reliable supervision, Flow4D successfully learns to perform temporal reasoning without ground-truth labels and achieves a respectable mean dynamic normalized error of 0.330.
Applying the same framework to the more advanced $\Delta$Flow backbone yields a performance boost on all categories, reducing the overall dynamic error by 19.7\% (0.330 to 0.265) and three-way EPE by 22.3\% (5.70 to 4.43).
This improvement aligns with the supervised results~\cite{zhang2025deltaflow}, where $\Delta$Flow demonstrates stronger temporal representation and motion modeling than Flow4D.
Together, these experiments show that TeFlow is backbone-agnostic: it unlocks the self-supervised potential of existing multi-frame architectures (e.g., Flow4D and $\Delta$Flow) and can be seamlessly applied to future scene flow backbones as they emerge.

\begin{table*}[h]
\centering
\setlength{\tabcolsep}{0.8em}
\caption{
Ablation study on different multi-frame backbones within our self-supervised pipeline TeFlow.
Results are evaluated on the Argoverse 2 validation set with five input frames.
The results demonstrate that our self-supervised objective integrates seamlessly with different multi-frame backbones (Flow4D and $\Delta$Flow), showing that the method is architecture-agnostic and applicable to future multi-frame scene flow designs.
}
\vspace{-0.8em}
\label{tab:diff_backbone}
\resizebox{0.8\linewidth}{!}{
\begin{tabular}{l|ccccc|cccc} 
\toprule
\multicolumn{1}{c|}{\multirow{2}{*}{Backbone}} & \multicolumn{5}{c|}{Dynamic Bucket-Normalized ↓}                                   & \multicolumn{4}{c}{Three-way EPE (cm) ↓}      \\
\multicolumn{1}{c|}{}                                 & Mean           & CAR            & OTHER         & PED.           & VRU         & Mean          & FD             & FS   & BS    \\ 
\midrule
Flow4D                                      & 0.330   & 0.254  & 0.326  & 0.329  & 0.411  & 5.70  & 12.98 & 2.67 & 1.46  \\
$\Delta$Flow                                       & \textbf{0.265} & \textbf{0.198} & \textbf{0.275} & \textbf{0.295} & \textbf{0.293} & \textbf{4.43} & \textbf{10.36} & 1.86 & 1.08  \\
\bottomrule
\end{tabular}}
\end{table*}

\begin{table*}[h]
\centering
\setlength{\tabcolsep}{0.8em}
\caption{Ablation of the candidate voting and aggregation pipeline.
We evaluate three components: the directional consistency agreement matrix $\mathbf{M}$ in~\Cref{eq:direction}, 
the reliability weights $\mathbf{w}$ in~\Cref{eq:rel_weight}, and the aggregation step in~\Cref{eq:avg_consesus_flow}. 
Removing either directional consistency or reliability weighting degrades performance, 
with $\mathbf{M}$ playing the dominant role in stabilizing candidate votes.
Skipping aggregation further reduces accuracy by discarding supportive consistent candidates.
Results on the Argoverse~2 validation set demonstrate that the full pipeline 
is essential for producing reliable multi-frame supervision.
}
\vspace{-0.8em}
\label{tab:ab_voting}
\resizebox{\linewidth}{!}{
\begin{tabular}{c|l|ccccc|cccc} 
\toprule
\multicolumn{1}{c|}{\multirow{2}{*}{Exp. Id}} & \multicolumn{1}{l|}{\multirow{2}{*}{Ablation Variant}} & 
\multicolumn{5}{c|}{Dynamic Bucket-Normalized ↓} & 
\multicolumn{4}{c}{Three-way EPE (cm) ↓} \\
& & Mean & CAR & OTHER & PED. & VRU 
                     & Mean & FD & FS & BS \\ 
\midrule
1 & w/o $\mathbf{M}$ (i.e., $\mathbf{M}=\mathbf{1}$)   
& 0.349 & 0.301 & 0.408 & 0.292 & 0.394  
& 5.81 & 15.33 & 1.34 & 0.74 \\
2 & w/o $\mathbf{w}$ (i.e., $\mathbf{w}=\mathbf{1}$)   
& 0.271 & 0.199 & 0.300 & \textbf{0.285} & 0.302  
& 5.70 & 12.98 & 2.67 & 1.46 \\
3 & w/o Aggregation                                    
& 0.289 & 0.226 & 0.332 & 0.303 & 0.295  
& 4.61 & 11.55 & 1.47 & 0.82 \\
4 & TeFlow (Ours)
& \textbf{0.265} & \textbf{0.198} & \textbf{0.275} & 0.295 & \textbf{0.293}  
& \textbf{4.43} & \textbf{10.36} & 1.86 & 1.08 \\
\bottomrule
\end{tabular}}
\vspace{-1em}
\end{table*}

\paragraph{Ablation Study on Candidate Voting and Aggregation}
\change{As detailed in~\Cref{sec:tem_ensem}, our consensus formulation operates in two stages: (a) voting, which scores candidates based on directional consistency agreement $\mathbf{M}$ and reliability weights $\mathbf{w}$, and (b) aggregation, which averages all candidates consistent with the winner. We ablate these components to quantify their contribution to the supervisory signal.

\vspace{4pt}
\textit{1) Directional consistency agreement~\Cref{eq:direction}.}
The directional consistency agreement $\mathbf M$ is used to determine which candidates in the pool reinforce each other during voting.
We ablate it by replacing $\mathbf{M}$ with an all-ones matrix, ignoring agreement among candidates.
As shown in \Cref{tab:ab_voting} (Experiment 1), removing directional consistency leads to a notable performance drop: the dynamic bucket-normalized error increases from 0.265 to 0.349, corresponding to a 31.7\% relative degradation. 
This demonstrates that directional agreement is important for filtering out inconsistent motion hypotheses and stabilizing the temporal ensembling process.
The only exception is pedestrians, where performance slightly decreases. 
Their motion frequently changes direction within short windows, so enforcing strict agreement can suppress valid short-term cues.

\vspace{4pt}
\textit{2) Reliability weighting~\Cref{eq:rel_weight}.}
The reliability weights $\mathbf{w}$ are used to emphasize candidates that provide clearer motion cues.
We ablate this by setting all weights to one, removing both magnitude-based emphasis and temporal decay.
As shown in \Cref{tab:ab_voting} (Experiment 2), this leads to a moderate increase in dynamic bucket-normalized error (0.265 to 0.271).
The impact is particularly evident in absolute accuracy, where the mean Three-way EPE degrades significantly from 4.43 cm to 5.70 cm (a 28.7\% error hike).
This demonstrates that reliability weighting serves as a refinement for precision: while directional consistency filters outliers, weighting refines the consensus by emphasizing stronger motion cues.
Similar to the directional-consistency ablation (Experiment 1), pedestrians exhibit a slightly different trend, suggesting that strictly favoring larger motions may occasionally suppress short-term but valid cues for rapidly changing agents.

\vspace{4pt}
\textit{3) Flow aggregation~\Cref{eq:avg_consesus_flow}.}
In our flow aggregation, we use not only the consensus winner but also supporting candidates that agree with it.
We ablate this aggregation step by supervising with the consensus winner alone, i.e., $\bar{\mathbf f}_{\mathcal{C}_j}=\mathbf f_{a^{\dagger}}$.
As shown in \Cref{tab:ab_voting} (Experiment 3), removing flow aggregation leads to a drop in accuracy: the dynamic bucket-normalized error increases from 0.265 to 0.289, a 9.1\% relative increase. 
Although the consensus winner achieves the highest vote score, it remains a single estimation sample. Consequently, relying on it directly leaves the supervision susceptible to the specific noise or quantization artifacts of that individual candidate.
In contrast, aggregating all directionally consistent flows reinforces stable temporal evidence and suppresses spurious single-frame deviations, leading to smoother supervision and consistently better overall accuracy.

\vspace{4pt}
Together, these ablations show that our proposed voting pipeline is necessary for producing stable and reliable multi-frame supervision. 
Directional consistency filters contradictions, reliability weights prioritize trustworthy cues, and aggregation consolidates evidence to mitigate isolated noise. 
Combining all three components yields the best self-supervised training performance by providing a reliable supervisory signal.
}

\paragraph{Analysis on Hyperparameter Selection}
We further analyze the sensitivity of TeFlow to its key hyperparameters by varying one parameter at a time while keeping the others fixed at their optimal values. Results are reported in~\Cref{tab:ab_param} and provide additional insight into the functioning of the temporal ensembling strategy.

\begin{table*}[ht]
\centering
\setlength{\tabcolsep}{0.8em}
\caption{
Ablation study on the key hyperparameters of TeFlow, evaluated on the Argoverse 2 validation set. 
The default and best-performing configuration is cosine similarity $\tau_{cos}=0.7$ (45°), Top-K = 5, and time decay $\gamma=0.9$. 
In each row, only the specified parameter is varied from this setting.
}
\label{tab:ab_param}
\vspace{-0.8em}
\resizebox{0.86\linewidth}{!}{
\begin{tabular}{l|ccccc|cccc} 
\toprule
\multicolumn{1}{c|}{\multirow{2}{*}{TeFlow Setting}} & \multicolumn{5}{c|}{Dynamic Bucket-Normalized ↓}                          & \multicolumn{4}{c}{Three-way EPE (cm) ↓}      \\
\multicolumn{1}{c|}{}                                & Mean           & CAR            & OTHER & PED.           & VRU            & Mean          & FD             & FS   & BS    \\ 
\midrule
Default                                              & \textbf{0.265} & \textbf{0.198} & \textbf{0.275} & 0.295          & \textbf{0.293} & \textbf{4.43} & \textbf{10.36} & 1.86 & 1.08  \\
$\tau_{cos}=0$ (90°)                                    & 0.307          & 0.239          & 0.365 & 0.291          & 0.332          & 5.19          & 13.02          & 1.60 & 0.95  \\
$\tau_{cos}=0.9$ (20°)                               & 0.289          & 0.207          & 0.356 & 0.294          & 0.297          & 4.42          & 10.41          & 1.80 & 1.04  \\
$K=20$                                               & 0.353          & 0.283          & 0.355 & 0.312          & 0.463          & 5.88          & 14.97          & 1.68 & 1.00  \\
$K=10$                                               & 0.307          & 0.241          & 0.314 & 0.296          & 0.377          & 5.11          & 12.39          & 1.83 & 1.12  \\
$\gamma=1$                                            & 0.303          & 0.224          & 0.348 & 0.311          & 0.330          & 4.73          & 11.55          & 1.66 & 0.98  \\
$\gamma=0.5$                                          & 0.285          & 0.232          & 0.308 & \textbf{0.290} & 0.311          & 4.92          & 11.65          & 1.98 & 1.12  \\
\bottomrule
\end{tabular}
}
\end{table*}

\vspace{4pt}
\noindent1) \textit{Top-K} This parameter controls the number of external candidates in the candidate pool. A small, high-quality set proves most effective, with the best performance at $K=5$. Larger values introduce noise from less reliable geometric matches and degrade accuracy.

\vspace{4pt}
\noindent2) \textit{Cosine Similarity} This threshold determines which candidates are included in the consensus matrix. The optimal value of 0.707 (45°) strikes the right balance: looser thresholds allow inconsistent motions, while stricter ones discard valid candidates too early.

\vspace{4pt}
\noindent3) \textit{Time Decay} This factor weights candidates by their temporal distance, giving higher importance to recent frames. Our default of $\gamma=0.9$ outperforms both no decay ($\gamma=1.0$) and stronger decay ($\gamma=0.5$). Without decay, distant frames are treated equally and introduce noise, while overly strong decay underutilizes longer-term consistency that benefits large, predictably moving objects.

\begin{figure*}[h]
\centering
\includegraphics[width=\linewidth]{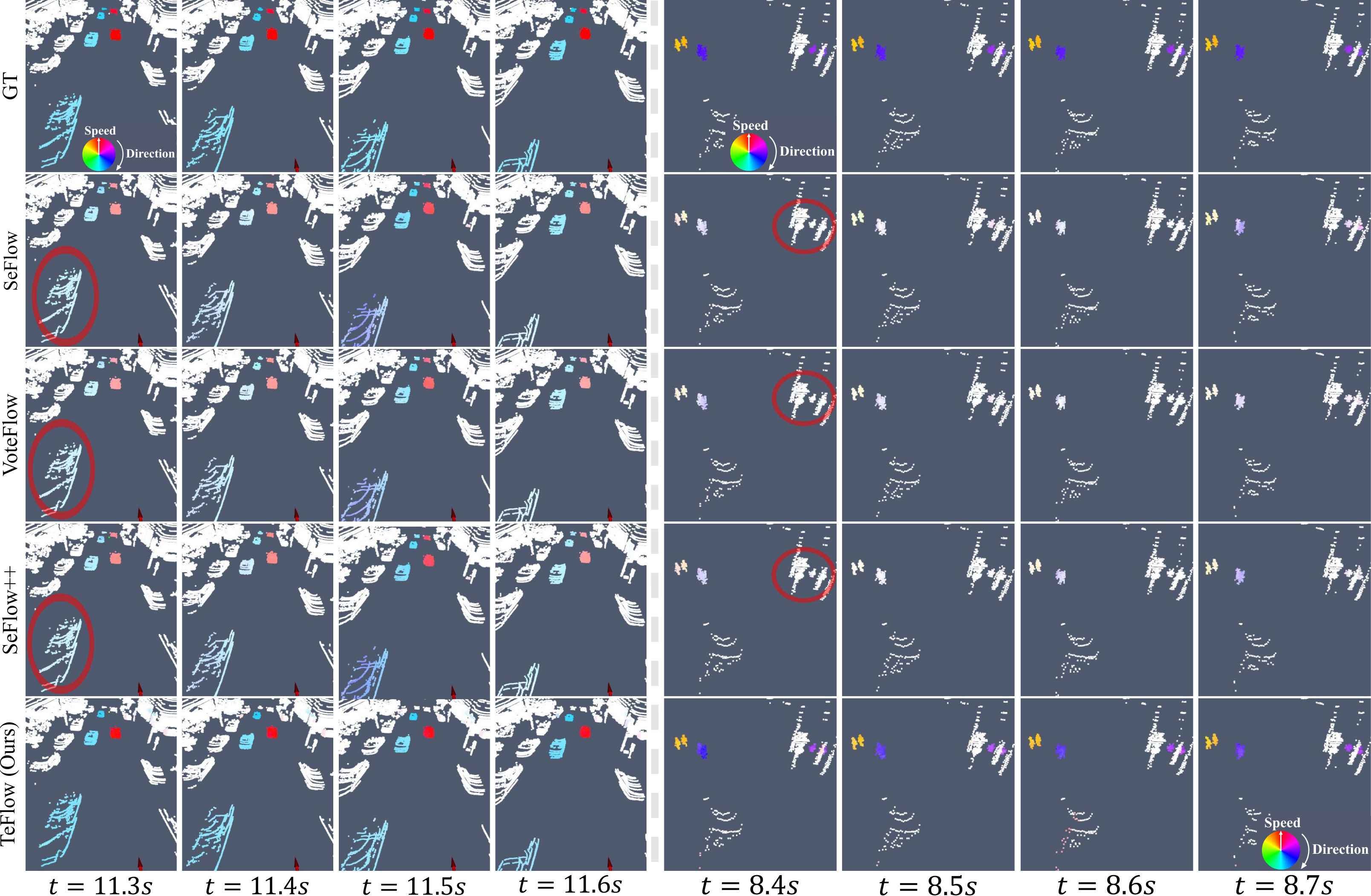}
\vspace{-1.5em}
\caption{
Qualitative comparisons on the Argoverse 2 validation set. Left: A multi-vehicle scene. Right: A vehicle stopping for pedestrians. Our method robustly handles both scenarios, unlike the baseline. (Best viewed in color.)
The scenes correspond to scene IDs `c85a88a8-c916-30a7-923c-0c66bd3ebbd3' and `b6500255-eba3-3f77-acfd-626c07aa8621'.
}
\label{fig:qual_two_av2}
\end{figure*}
\section{Qualitative Results}
\label{app:qual}
The qualitative results in the main paper are derived from the scenes `8749f79f-a30b-3c3f-8a44-dbfa682bbef1' and `scene-0104' in the Argoverse 2 and nuScenes validation set, respectively. 

Here, we present additional qualitative results comparing our TeFlow with top self-supervised feed-forward methods, namely SeFlow~\cite{zhang2024seflow}, VoteFlow~\cite{lin2025voteflow}, and SeFlow++~\cite{zhang2025himo}. 
All visualizations use a standard color-coding scheme, where hue indicates motion direction and saturation encodes speed.

\begin{figure*}[h]
\centering
\includegraphics[width=\linewidth]{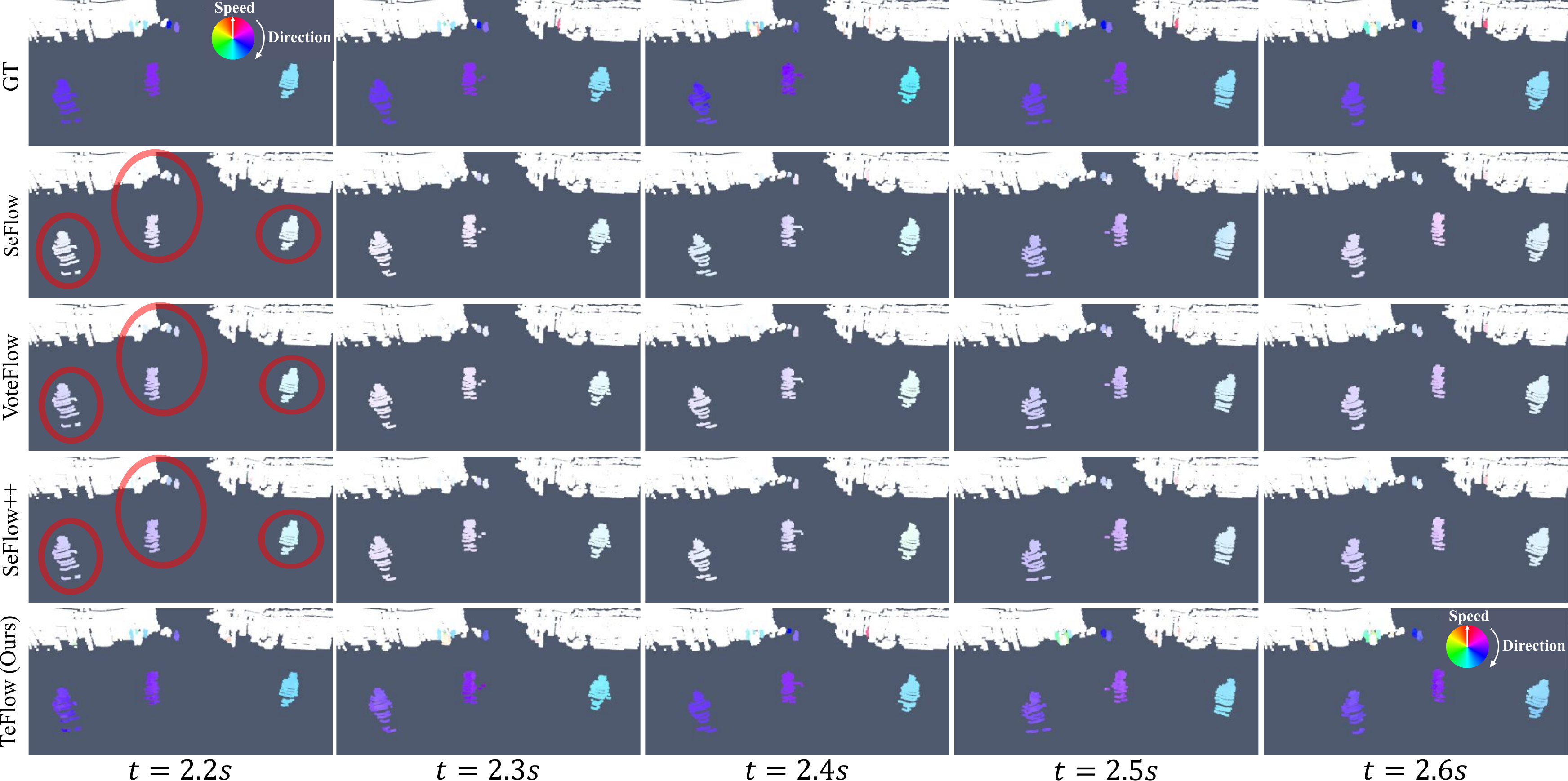}
\vspace{-1.5em}
\caption{
Qualitative results on the Argoverse 2 validation set. Our method accurately captures the motion of multiple pedestrians, while all feed-forward baselines underestimate the flows of moving pedestrians. (Best viewed in color.)
The scenes correspond to scene IDs `9f871fb4-3b8e-34b3-9161-ed961e71a6da'.
}
\vspace{-1em}
\label{fig:qual_three_people}
\end{figure*}
\begin{figure*}[t]
\centering
\includegraphics[width=\linewidth]{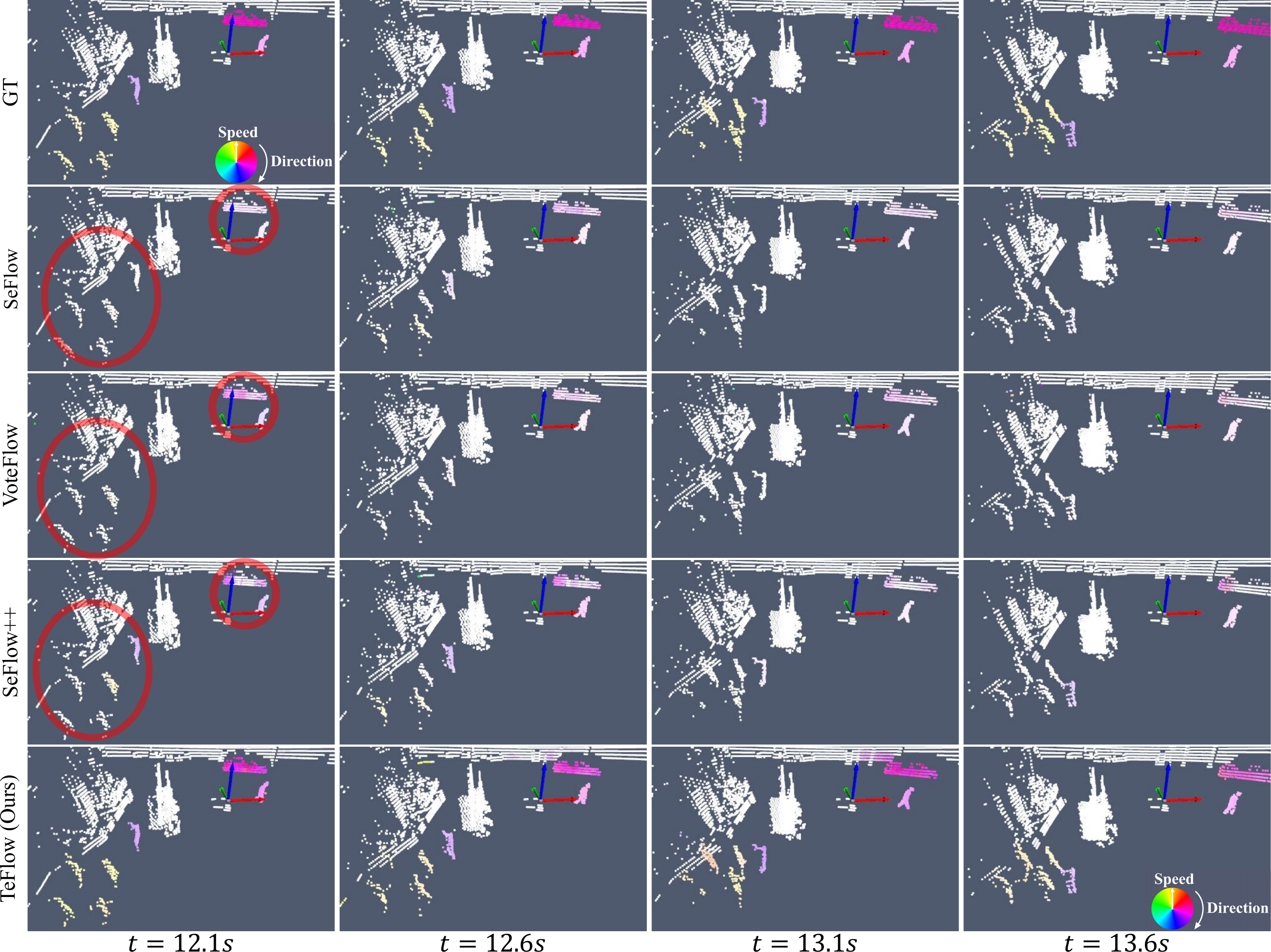}
\caption{
Qualitative results on the nuScenes validation set. On this sparser data, TeFlow provides complete motion for the vehicle and detects the pedestrians, whereas the baseline underestimates the car's flow and misses the smaller actors. (Best viewed in color.)
The scenes correspond to the scene IDs `scene-0025'.
}
\vspace{-1em}
\label{fig:qual_nus}
\end{figure*}
\begin{figure*}[t]
\centering
\includegraphics[width=\linewidth]{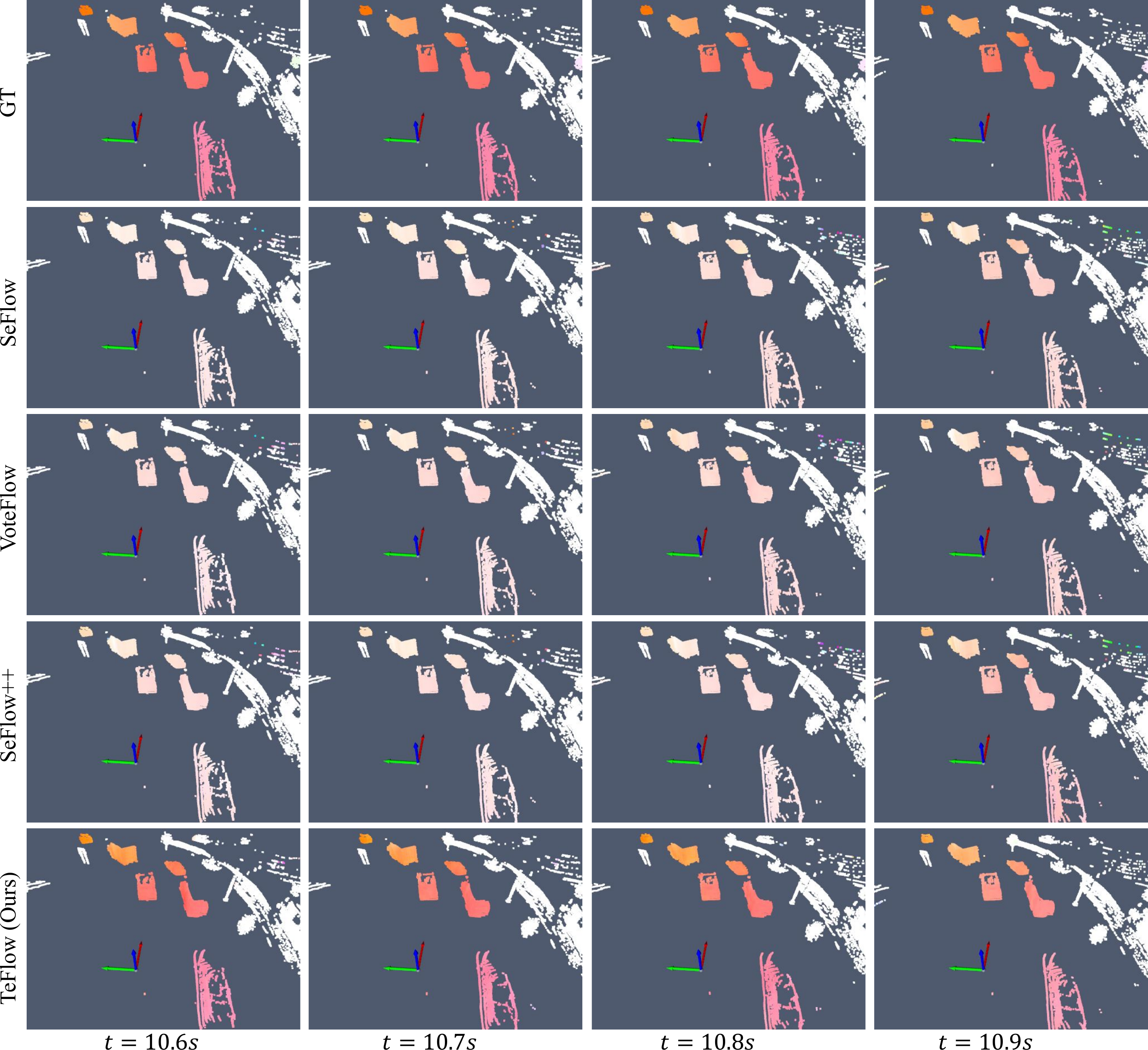}
\caption{
Qualitative results on the Argoverse 2 validation set. Our method accurately captures the motion of vehicles in complex roundabout scenarios. (Best viewed in color.)
The scenes correspond to scene IDs `bdb9d309-f14b-3ff6-ad1f-5d3f3f95a13e'.
}
\vspace{-1em}
\label{fig:qual_round}
\end{figure*}

\Cref{fig:qual_two_av2} shows two complex multi-agent scenes from Argoverse 2. In the left scene, three oncoming vehicles are captured. While the ground truth indicates consistent forward motion, all baseline feed-forward methods occasionally predict conflicting directions (e.g., flows shift from blue to purple around $t=11.3$–$11.5$), reflecting the instability of two-frame supervision. In contrast, our TeFlow maintains coherent motion across time, producing stable and accurate flow for each vehicle.
The right scene highlights another common failure case: pedestrians motion. The ground truth reveals clear trajectories, including a distant pedestrian partially occluded by a lamp post. Baseline methods consistently underestimate the flow magnitudes of these small or occluded agents, resulting in weak or inconsistent predictions. While our TeFlow captures their motion with the correct magnitude and direction.

\Cref{fig:qual_three_people} presents a challenging scene with three pedestrians crossing the road simultaneously. In the ground truth, all pedestrians exhibit clear motion, yet baseline feed-forward methods underestimate their flow magnitudes due to noisy two-frame supervision, resulting in weak and inconsistent predictions under such dynamic motion. In contrast, the model trained with our TeFlow objective produces flow fields that are both spatially coherent and temporally stable. Each motion of pedestrian is captured with accurate magnitude and direction, closely matching the ground truth across the time window. Furthermore, TeFlow also preserves reliable estimates for other small or distant dynamic objects, highlighting its robustness under challenging scenarios with sparse observations.

\Cref{fig:qual_nus} shows a challenging scene from the nuScenes validation set. In the lower-left corner, five pedestrians are walking together, while a vehicle and another pedestrian are passing in front of the ego car. The ground truth indicates clear motion for both the vehicle and pedestrians. However, baseline feed-forward methods significantly underestimate the vehicle’s flow magnitude and often fail to detect the motions of the smaller pedestrians. In contrast, TeFlow produces a smooth and complete flow field for the vehicle and successfully captures the individual motions of the pedestrians, even under the sparse point density of nuScenes.

\Cref{fig:qual_round} illustrates a complex roundabout scene from the Argoverse 2 validation set. Multiple vehicles are moving along curved trajectories. The baseline methods fail to provide consistent estimates, often underestimating the motion or producing fragmented flows, especially for vehicles entering or exiting the roundabout. While, TeFlow produces coherent and smooth flow fields that closely follow the ground-truth directions, demonstrating its ability to handle complex multi-agent interactions in curved motion scenarios.

\end{document}